\documentclass[sigconf, nonacm]{acmart}
\usepackage{subcaption}
\usepackage{graphicx}
\usepackage{hyperref}
\AtBeginDocument{%
  \providecommand\BibTeX{{%
    \normalfont B\kern-0.5em{\scshape i\kern-0.25em b}\kern-0.8em\TeX}}}

\setcopyright{acmcopyright}
\copyrightyear{2022}
\acmYear{2022}
\acmDOI{XXXXXXX.XXXXXXX}

\acmConference[Pre-print]{Make sure to enter the correct
  conference title from your rights confirmation emai}{June 03--05,
  2018}{Woodstock, NY}
\begin{document}

\title{City-Wide Perceptions of Neighbourhood Quality using Street View Images}

\author{Emily Muller}
\affiliation{%
  \institution{Imperial College London}
  \country{UK}
}
  \email{emily.muller@imperial.ac.uk}

\author{Emily Gemmell}
\affiliation{%
  \institution{University of British Columbia}
  \city{Vancouver}
  \country{Canada}}
\email{emily.gemmell@ubc.ca}

\author{Ishmam Choudhury}
\affiliation{%
  \institution{University of British Columbia}
  \city{Vancouver}
  \country{Canada}}
\email{ishmam1998@gmail.com}

\author{Ricky Nathvani}
\affiliation{%
  \institution{Imperial College London}
  \country{UK}}
\email{r.nathvani@imperial.ac.uk}

\author{Antje Barbara Metzler}
\affiliation{%
  \institution{Imperial College London}
  \country{UK}}
\email{antje.metzler18@imperial.ac.uk}

\author{James Bennett}
\affiliation{%
  \institution{Imperial College London}
  \country{UK}}
\email{umahx99@imperial.ac.uk}

\author{Emily Denton}
\affiliation{%
  \institution{Google}
    \city{New York}
  \country{USA}}
\email{dentone@google.com}

\author{Seth Flaxman}
\affiliation{%
  \institution{University of Oxford}
  \country{UK}}
\email{seth.flaxman@cs.ox.ac.uk}

\author{Majid Ezzati}
\affiliation{%
  \institution{Imperial College London}
  \country{UK}}
\email{majid.ezzati@imperial.ac.uk}







\renewcommand{\shortauthors}{Muller et al.}

\begin{abstract}
The interactions of individuals with city neighbourhoods is determined, in part, by the perceived quality of urban environments. Perceived neighbourhood quality is a core component of urban vitality, influencing social cohesion, sense of community, safety, activity and mental health of residents. Large-scale assessment of perceptions of neighbourhood quality was pioneered by the Place Pulse projects. Researchers demonstrated the efficacy of crowd-sourcing perception ratings of image pairs across 56 cities and training a model to predict perceptions from street-view images. Variation across cities may limit Place Pulse’s usefulness for assessing within-city perceptions. In this paper, we set forth a protocol for city-specific dataset collection for the perception: `On which street would you prefer to walk?’. This paper describes our methodology, based in London, including collection of images and ratings, web development, model training and mapping. Assessment of within-city perceptions of neighbourhoods can identify inequities, inform planning priorities, and identify temporal dynamics. Code available: \url{https://emilymuller1991.github.io/urban-perceptions/}.
\end{abstract}



\begin{teaserfigure}
\centering
  \includegraphics[width=1\textwidth]{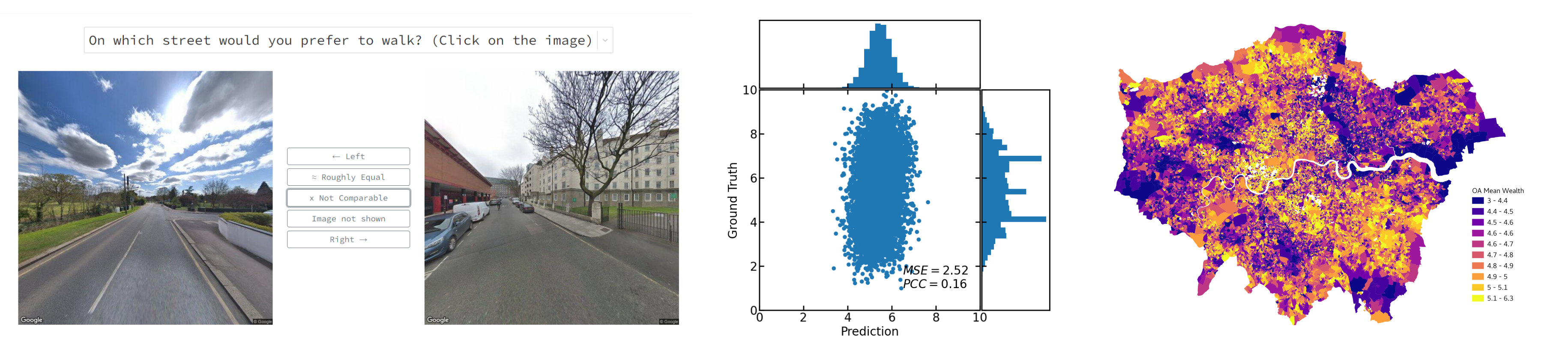}
  \caption{(Left) Web application for user survey. (Middle) Deep Learning model prediction histogram. (Right) Inference in London.}
  \Description{3 Figures arranged in a row. (Left) The web app display shown to users. (Middle) Histogram of predicted walk scores versus ground truth for test data set. (Right) Map of Greater London with Output Areas coloured by average walk score decile.}
  \label{fig:teaser}
\end{teaserfigure}

\maketitle

\section{Introduction}
Street quality refers to the overall quality of the street environment, including built and natural environments, human interactions, infrastructure and transport modes \cite{ibrahim2020understanding}. Street quality is relevant for various components of city life such as urban mobility, safety, social cohesion and economic productivity \cite{jacobs1961jane}. The call for a renewed livelihood of street space centres around human-oriented design \cite{gehl2013cities}, a critical perspective to enhance urban vitality and alleviate the environmental burdens faced by urban residents. UK initiatives to achieve these objectives include Better Streets London \cite{better_streets} and Low Traffic Neighbourhoods \cite{ltn}. Within urban design, visual measures of urban physicality contribute to street quality, such as greenery, enclosures, openness, street wall continuity, accessibility, and scale \cite{tang2019measuring}. Objective measures of street quality have been defined by combining these or other attributes to produce city scores \cite{walk_score, tranquil_score}. This approach, however, does not reveal the perceived quality of the street as experienced by the user. When viewed through the lens of environmental psychology, perception provides a meaningful interpretation for social behaviour \cite{steg2013environmental}. For example, streets which were perceived as being safe were found to have lower associated crime rates \cite{ordonez2014learning}. As such, perceptions of street quality goes beyond the aesthetic realm, often viewed as qualitative and idiosyncratic, to well defined measures that impact user experience. By integrating user perceptions in urban planning, we can inform design approaches which promote the health of urban residents.

The \textit{Image of the City}, as formalised by Kevin Lynch \cite{lynch1964image}, is `the product of both immediate sensation and the memory of past experience... each individual creates and bears their own image, but there seems agreement amongst members of the same group'. While Lynch refers to the mental image, photographic images have long been used to conduct visual surveys of street quality \cite{rundle_using_2011, griew_developing_2013, gebel2009correlates}. The ability to scale these measurements city-wide has become possible only in the last decade, due to increased urban surveillance in the form of street view and satellite imagery, and the accessibility of such data. Crowd-sourced pairwise image ratings have been used to collect perceptions on safety \cite{salesses_collaborative_2013}, beauty \cite{quercia_aesthetic_2014}, scenic-ness \cite{seresinhe2015quantifying} and overall street quality \cite{tang2019measuring}. The large scale global project called Place Pulse 2.0 \cite{dubey_deep_2016} has provided a dataset and methodology for measuring urban perceptions which has been adopted further to new and unseen cities \cite{zhang2018measuring, de2016safer, ordonez2014learning, porzi2015predicting}. This dataset provides an extremely rich basis for generalisability, owing to its surveying across $56$ countries and collecting over $1$ million user perceptions globally. The trade-off for generalisability is that city-specific perceptions are diminished. Modern city street view images are the unique snapshots of historical, cultural and planning processes, and what may be perceived as apparently wealthy or safe in one city, does not necessarily translate to another city or the image held of wealth or safety for those citizens. Therefore, this work examines within-city perceptions of street quality by including images from only London. In this paper, we set forth a protocol for city-specific urban perception collection using the exemplar question `On which street would you prefer to walk?’. We test our method in London, UK. This paper describes our methodology, including web development for hosting image-based survey, data collection and deep learning model training. We present results and interpretations of model predictions, with mapping across London neighbourhoods.

\section{Background}
Much of the work within the domain of street view images for measuring the urban environment takes a supervised learning approach, in which area-level outcomes, such as census data, are assigned to images and models are then trained to predict the outcome assignment. Notable studies use street view images to predict socio-economic status \cite{suel2019measuring}, air pollution \cite{suel2022you}, house prices \cite{law2019take} and urban greenery \cite{li2015assessing}. Other studies have attempted to increase interpretability by further image analysis. For example, vehicle types were extracted from street view images and found to predict voting preferences \cite{gebru_using_2017}, semantic pixel-wise labels (such as sky, car, road, sidewalk) were extracted from images to predict air pollution \cite{qi2021using} and metrics of blue and green space from images were found to be associated with geriatric depression \cite{helbich_using_2019}. While this work has received much attention, important limitations of supervised learning remain. That is, assignment of area-level attributes to image-level areas can smooth over interesting heterogeneity within a region which images capture on smaller scales. Furthermore, outcome labels may be scarce or unavailable. In the absence of image-level labels, one approach is to perform unsupervised learning. Seminal work using unsupervised learning clusters extracted features from images to identify architectural typologies of Paris \cite{doersch2012makes}. Another approach employed similarity scores to build a platform where users can query aspects of the environment, such as the entry staircase of a New York Brownstone, and multiple similar images are returned along with their geo-tagged locations \cite{miranda_urban_2020}. The latter example is especially interesting as it demonstrates use of a human-in-the-loop interface which can provide the validation difficult to accomplish in unsupervised learning. Image-level labeling can also be achieved using pre-trained models such as Places365 which predicts scenes such as highway, park and slum \cite{zhou2017places}. This is, however, constrained by the data sets and methods which are available for extracting features. 

Data set creation provides the flexibility to label images in a way that is relevant to a given substantive research domain. This can be done by labeling a subset of images and inferring the predicted outcome classes on a city-wide scale, as in \cite{liu2017machine}. A more generalisable approach collects ratings of image pairs using crowd-sourcing \cite{salesses_collaborative_2013, quercia_aesthetic_2014,seresinhe2015quantifying, tang2019measuring}. In this scenario, individuals are elicited to provide their perceptions by being shown an image pair side by side and asked to vote on one of the images. What is gained by casting the net of users wide, is a generalisable model of how individuals experience the city with respects to a given perception, and a meaningful proxy for social activity. For example, higher perceptions of safety were shown to be associated with more lively neighbourhoods for certain population groups \cite{de2016safer}, and better mental health and well-being \cite{won2016neighborhood}, and perceptions of playability were negatively associated with high crime and traffic \cite{kruse2021places}. Discordance between objective and perceptive measures of the urban environment \cite{gebel2009correlates} can provide new information to guide sustainable city development. 

In this work, we outline a methodology that enables researchers to collect city-specific image-level perception scores which can be used to provide proxy information on urban activity at small spatial scales. We analyse demographic differences in perceptive preferences, train a deep convolutional neural network to predict overall perception scores on a hold out test set, compare Place Pulse 2.0 model accuracy to the new perception of walkability. We then infer walkability and Place Pulse 2.0 perception scores on over 1 million London images, providing maps at small area-levels for each perception, and examine features which correspond to high and low perception scores.

\section{Methodology}
\subsection{Images}
This study uses images collected using the Google Street View API \cite{google}, an online image service with good street coverage in London. Metadata requests have as input latitude and longitude (\textit{lat, lon}) pairs and return all images across all years found close to that point. We produced a sequence of input (\textit{lat, lon}) pairs at 20 meter intervals arranged on a square grid across the Greater London Authority spatial extent \cite{statistics_2019}. When metadata was collected in 2019, the Street View API returned a total of 1,095,752 images taken in 2018 on Greater London streets. Points are sampled at 20 meter intervals on the London Street Network \cite{open_roads} resulting in 934,791 point locations. Images are then sampled based on nearest distance to road point. The assumption is that this covers a contiguous visual representation of the street without repetition. The final database totals 633,419 unique images, covering $67\%$ of London roads. Two images per street view location are selected perpendicular to the street to capture the building front, as in \cite{law_street-frontage-net_2020}. This is done in \textit{Python 3.7} \cite{python} using the Unix software \textit{gnu parallel} \cite{Tange2011a} to utilise CPU cores for fast implementation, Python package \textit{robolyst/streetview} \cite{robolyst} to query Google Street View API, \textit{Qgis 3.27} \cite{QGIS_software} and \textit{postgres} \cite{postgres} for geo-spatial processing. Code is available \href{https://github.com/emilymuller1991/urban-perceptions/tree/main/download_images}{here}.

\subsection{Survey design} 
We collected user-rated perceptions using an online, crowd-sourcing approach \cite{dubey_deep_2016, quercia_aesthetic_2014, tang2019measuring}. Users were shown two images side by side and asked to make a binary decision: `On which street would you prefer to walk?'. This differs from studies which explore gradients of scores \cite{liu2017machine, seresinhe2015quantifying}. While these do offer more information, the additional time-cost impedes the scalability of crowd-sourcing. Accurately aggregating the information from these binary image tasks relies on the number of times that images are compared across the dataset, and the agreement between sets of pairs. Since our image dataset is very large, we take a stratified sample of $\sim 25$K images to be ranked. This reduces the required number of ratings (and raters) while ensuring we can train a deep learning model \cite{goodfellow2016deep}. We stratify images based on similar groups of built environment characteristics such as vegetation, high-density, commercial and residential areas. Groups are defined using R-MAC feature extraction \cite{tolias_particular_2016} followed by k-means clustering. Images pairs are shown with a weighting towards within group comparisons to better distinguish perceptions of similar streets. This is achieved by adding a mixing coefficient to sample the image pairs. The result is that on average, $20\%$ of image pairs are within the same cluster.

\subsection{Implementation}
We have used Flask \cite{grinberg2018flask}, a micro web framework written in Python, for the back-end. This uses the module \textit{psycopg} \cite{copg2} to connect to the postgres database which hosts image metadata, user perceptions and user demographics. We developed the front-end using React \cite{react}, a JavaScript library, which hosts the user interface as well as making GET and POST requests to the back-end.

We deployed the web app using a Kubernetes \cite{Kubernetes}, after containerising the front and back end in Docker \cite{merkel2014docker}. We established an endpoint for the postgres database to connect to the DigitalOcean droplet where the database is stored. Code is available \href{https://github.com/emilymuller1991/urban-perceptions/tree/main/web_app}{here}.

Data collection ran for two months from the end of April 2022 until June 2022, after receiving ethics approval. We used personal networks to disseminate the website, including mailing groups and social media. We used Amazon Mechanical Turk (AMT) to collect additional user-rated perceptions ensuring that workers were paid above minimum wage. In addition, we excluded workers which did not have a \textit{Masters} qualification rating as given by the platform. 

\section{Survey Results}

\subsection{Participant demographics}
A total of 25,154 images were included in the database. Basic statistics for the image ratings are shown in Table \ref{tab:demo}. Of the total pairwise ratings, around 16\% were recorded as not shown --- the image is not loaded from the Google server. This could be because its unique address has been changed, or, an error exists in the API key. $1$K images were rated as not comparable by the user. We removed all users who had greater than 90\% bias to choosing one side (left or right) over the other, and removed games which were played by the same user within the same 1 minute interval. This indicated duplicate entries in the postgres database. A total of 25,987 ratings are included in the final analysis. Of those games, 4,040 were collected using personal network crowd-sourcing and the remaining 21,947 were collected using AMT. Of the $207$ total individual raters, $59$ are non-AMT and $148$ are AMT workers. More than 86\% of individuals provided their demographic information.

\begin{table}
  \small 
  \caption{Basic Descriptives}
  \begin{tabular}{lr}
    \toprule
    Pulse London Database & \\
    \midrule
    Images in database & 25,154 \\
    Pairwise ratings & 37,966 \\
    \quad Not comparable & 1,079  \\
    \quad Not shown & 5,842 \\
    \quad One-sided clicks & 3,379 \\
    \quad Duplicate choices & 1,241\\
    \textbf{Usable games} & \textbf{25,987} \\
    Users & 207 (180 demographic) \\
    Images per user & 132 mean \\
    Repeated image pairs (>10) & 14\\
    Games per repeated image pair & 60 mean \\
    Repeated pairs agreement& 62\%, 149 users \\
  \bottomrule
  \label{tab:demo}
\end{tabular}
\Description{Numbers are reported from survey collection including games played, number of people who have played.}
\hfill
\small
  \caption{Agreement within user groups for repeated image pairs, $\%$, and counts, $n$.}
  \begin{tabular}{lrrr}
    \toprule
    Group & Yes ($n$) & No ($n$) & Other ($n$)  \\
    \midrule
    AMT & $56\%$ $(127)$ & $63\%$ $(22)$ \\
    London & $68\%$ $(17)$ & $64\%$ $(83)$ & $61\%$ $(43)$ \\ 
    Female & $62\%$ $(47)$ & $66\%$ $(88)$ & $55\%$ $(8)$\\
    H. Activity & $60\%$ $(104)$ & $70\%$ $(39)$\\
  \bottomrule
  \label{tab:agree}
\end{tabular}
\Description{Numbers are reported as percentage of user who agreed on a majority image for each stratified group.}
\end{table}

\begin{figure}
 \begin{subfigure}{0.49\linewidth}
     \includegraphics[width=1\textwidth, trim=0 30 0 20,clip]{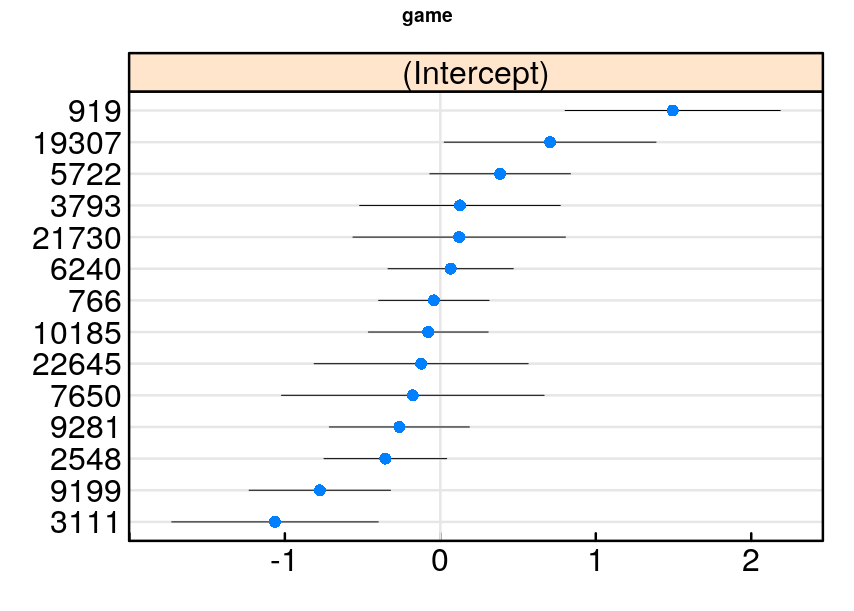}
     \caption{}
     \label{mlm:b}
 \end{subfigure}
  \begin{subfigure}{0.49\linewidth}
     \includegraphics[width=1\textwidth, trim=0 30 0 20,clip]{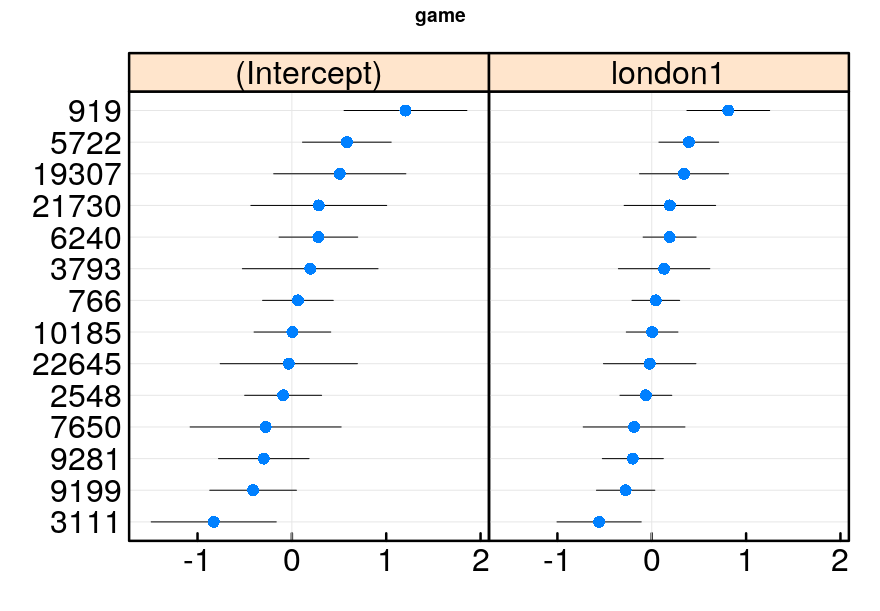}
     \caption{}
     \label{mlm:l}
 \end{subfigure}
 \begin{subfigure}{0.49\linewidth}
     \includegraphics[width=1\textwidth, trim=0 30 0 20,clip]{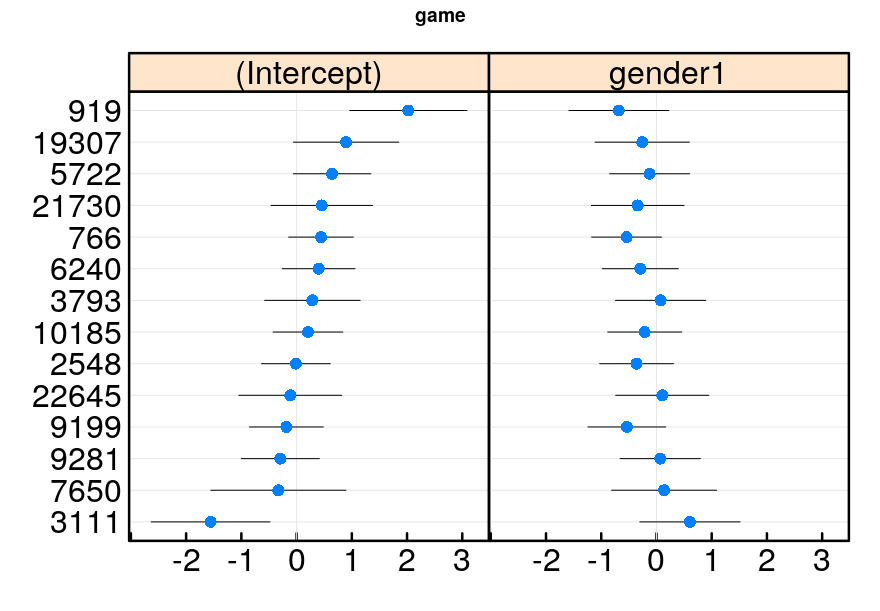}
     \caption{}
     \label{mlm:g}
 \end{subfigure}
  \begin{subfigure}{0.49\linewidth}
     \includegraphics[width=1\textwidth, trim=0 30 0 20,clip]{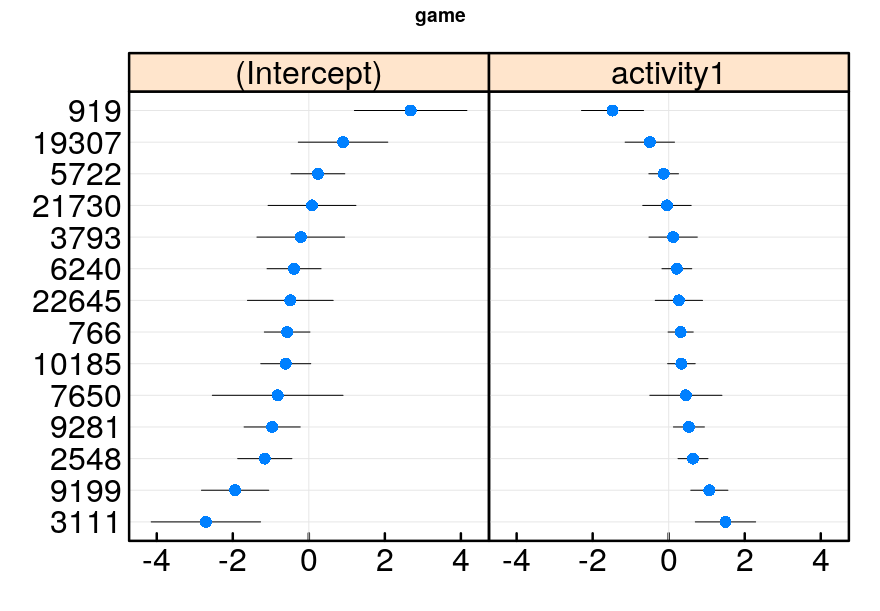}
     \caption{}
     \label{mlm:a}
 \end{subfigure}
 \caption{Multi-level model random effects and $95\%$ confidence intervals for image pairs (numbered on the y-axis).}
 \Description{4 plots showing image pairs on the y-axis and estimated random effects on the x-axis. The plot also shows 95\% confidence intervals across the x-axis, demonstrating those image pairs with a significant effect between groups. Top left shows baseline model with no group differences. Top right shows effects between London and not London. Bottom left shows effects between male and female. Bottom right shows effects between high activity and low activity.}
\end{figure}

\begin{figure}
 \begin{subfigure}[b]{1\linewidth}
     \includegraphics[width=0.49\textwidth]{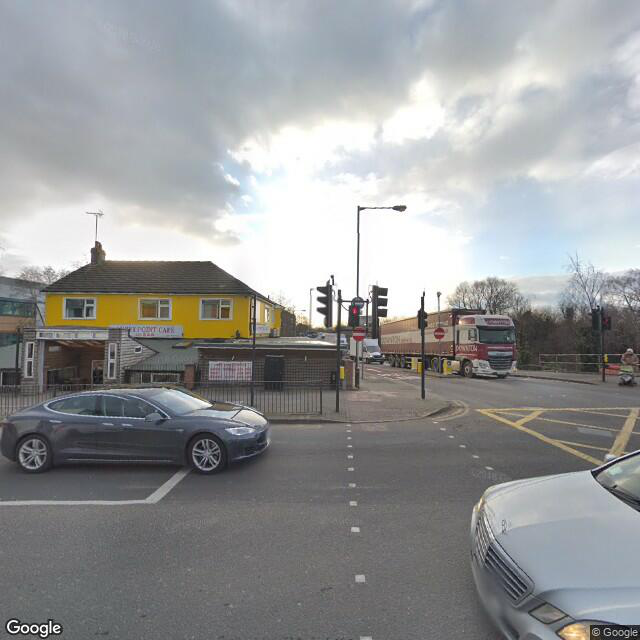}
     \includegraphics[width=0.49\textwidth]{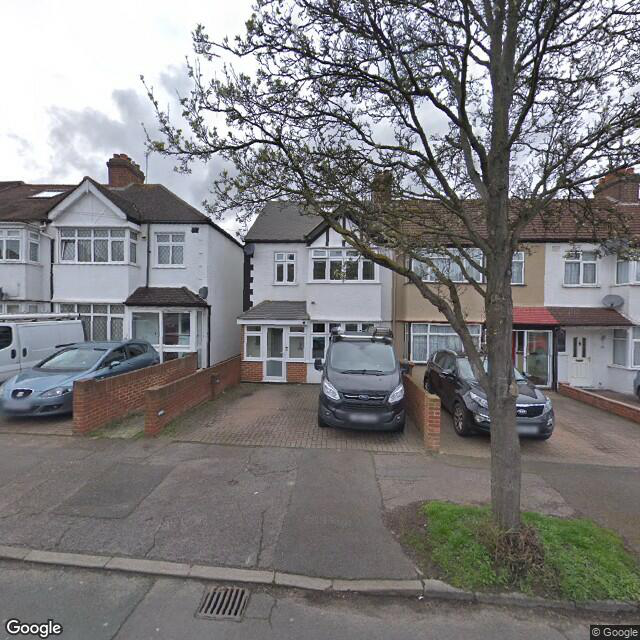}
     \caption{}
 \end{subfigure} 
 \begin{subfigure}[b]{1\linewidth}
     \includegraphics[width=0.49\textwidth]{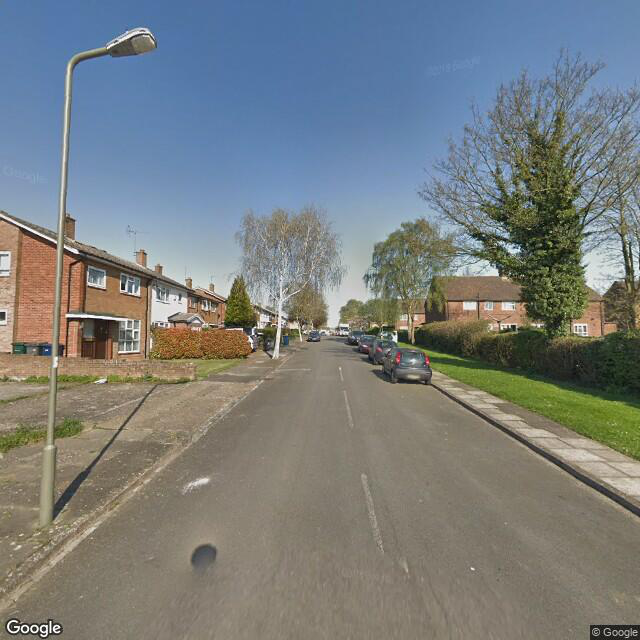}
     \includegraphics[width=0.49\textwidth]{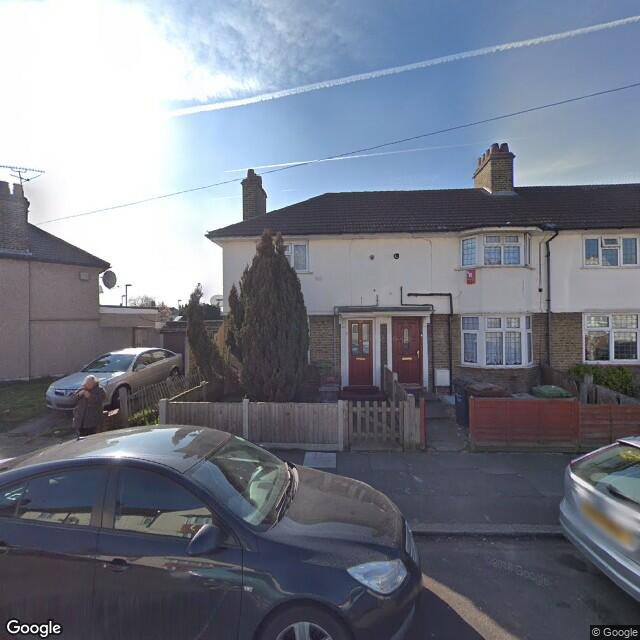}
     \caption{}
 \end{subfigure}
 \caption{Image pair 919 (a): for the baseline model with no groups, there is a significant effect for the right image choice with an $86\%$ consensus. Londoners are more likely to choose the right image than non-Londoners ($100\%$ vs $82\%$), after controlling for the baseline effect of the image pair. Individuals with high activity are more likely to choose the left image compared to low activity ($14\%$ vs $0\%$), after controlling for the baseline effect. Image pair 3111 (b): for the baseline model, there is a significant effect for the left image choice with $80\%$ consensus. Londoners are more likely to choose the left image, after controlling for the baseline effect of the image pair ($100\%$ vs $82\%$), whereas high activity individuals are more likely to choose the right ($100\%$ vs $0\%$).}
 \label{fig:sample_games}
 \Description{Top two images show the image pair 919 and bottom two images shows the image pair 3111.}
\end{figure}

\subsection{Quality assurance}
For the $14$ image pairs with more than $10$ repeated games we calculated the proportion of users who voted on the same image from the pair including only left and right votes (i.e., not equal). We chose the majority vote (i.e., $80\%$ and not $20\%$) to be the agreement between users and calculated the average across 14 image pairs. There is a $62\%$ agreement amongst $149$ users for one of the image pairs (see Table \ref{tab:demo}). For the same 14 image pairs, we stratified the raters into the following groups: AMT workers/non-AMT workers, London based/non-London based, female/male, high activity/low activity, where high activity is taking a walk at least twice a week. We observed the highest within group agreement for the low activity group, followed by the London group (see Table \ref{tab:agree}). 

We built a multi-level logistic regression model to test whether individuals from different groups (e.g. London vs.~not London) responded in significantly different ways to pairs of images. The model takes the form: $\pi_{ij}=\sigma(\beta_0 + u_{g|j})$ where for  $\pi_{ij}$ is the probability of individual $i$ choosing the right image of the pair of images $j$, i.e.~ the probability that $y_{ij}=1$. $u_{g|j} \sim \mathcal{N}(0,\sigma_u^2)$, are random effects for each group $g$, modelling the effect of being in group $g$ on the log-odds that $y_{ij}=1$ for image pair $j$. $\sigma$ is the standard logistic link function. For the baseline model without group differences, there are 4 image pairs which have a significant right or left leaning choice at a $95\%$ confidence interval (see Figure \ref{mlm:b}). Two of those image pairs (919 and 3111) are shown in Figure \ref{fig:sample_games}. For image pair 919, Londoners are more likely to choose the right image than non-Londoners ($100\%$ vs $82\%$), after accounting for the baseline effect of the image pair ($86\%$ choose right). Individuals with high activity are more likely to choose the left image compared to low activity ($14\%$ vs $0\%$), after accounting for the baseline effect of the image pair. For image pair 3111, Londoners are more likely to choose the left image ($100\%$ vs $82\%$), after accounting for the baseline effect of the image pair ($80\%$ choose right), whereas high activity individuals are more likely to choose the right ($100\%$ vs $0\%$).


\subsection{Ranking}
In order to rank individual images based on pairwise games, we implemented the Microsoft Trueskill algorithm \cite{graepel2007bayesian}. This algorithm initialises a score to each image from the normal distribution with $\mu=25$ and $\sigma=25/3$. $\mu$ is increased after a win and decreased after a loss. The extent of updates depends on each images' $\sigma$, i.e., how surprising the outcome is, given the image's existing score. These final scores were used to train the Convolutional Neural Network (CNN) in the next section.

\section{Deep CNN}
\begin{table*}[h!]
\small
  \caption{Resnet101 Model accuracy}
  \label{tab:cnn_results}
  \begin{tabular}{lrrrrrrr}
    \toprule
    Perception & Safety & Lively & Depressing & Boring & Beauty & Wealth & Walkability \\
        \midrule
    MSE test & 1.10 & 1.40 & 1.82 & 1.89 & 1.87 & 1.92 & 2.58 \\
    PCC$^*$ & 0.44 & 0.37 & 0.24 & 0.16 & 0.32 & 0.39 & 0.16\\
    Transfer MSE Test & 2.58  & 2.59 & 2.54 & 2.53 & 2.52 & 2.53 & - \\ 
    Transfer PCC & 0.14  & 0.15 & 0.16 & 0.13 & 0.16 & 0.16 & - \\ 
  \bottomrule
\end{tabular}

\footnotesize{$^*$ Pearsons Correlation Coefficient.}
\Description{Accuracies are reported for the deep models to predict perception scores (columns). Train, test and Pearsons correlations are reported (rows).}
\end{table*}

\begin{table*}[h!]
\small
  \caption{Model test errors and Games Multiplier=No. Games / No. Images}
  \label{tab:multiplier}
  \begin{tabular}{lrrrrrrr}
    \toprule
    Perception & Safety & Lively & Depressing & Boring & Beauty & Wealth & Walkability \\
        \midrule
    Games Multiplier & 4.65 & 3.34 & 2.01 & 1.59 & 1.36 & 1.31 & 1.03 \\
    Mean $\sigma$ TrueSkill & 3.43 & 3.95 & 4.95 & 5.45 & 5.71 & 5.73 & 6.00 \\
    MSE test & 1.10 & 1.40 & 1.82 & 1.89 & 1.87 & 1.92 & 2.52 \\
    PCC & 0.44 & 0.37 & 0.24 & 0.16 & 0.32 & 0.39 & 0.16\\
  \bottomrule
\end{tabular}
\Description{Accuracies are reported for the deep models to predict perception scores (columns). Train, test and Pearsons correlations, as well as the games multiplier, are reported (rows)}
\end{table*}

We separately trained deep CNN models across the original Place Pulse and the Pulse London datasets to predict perceptions of walkability. The Place Pulse dataset includes $110$K images and $6$ different outcome labels (perceptions of safety, beauty, liveliness, boring, depressing and wealth). The Pulse London dataset contains $25$K images with $1$ outcome label: walkability perception. We split both datasets into $65-5-30$, training, validation and test splits.

\begin{figure}[!ht]
\raisebox{5pt}{
\begin{subfigure}[b]{0.4\linewidth}
  \includegraphics[width=1\textwidth]{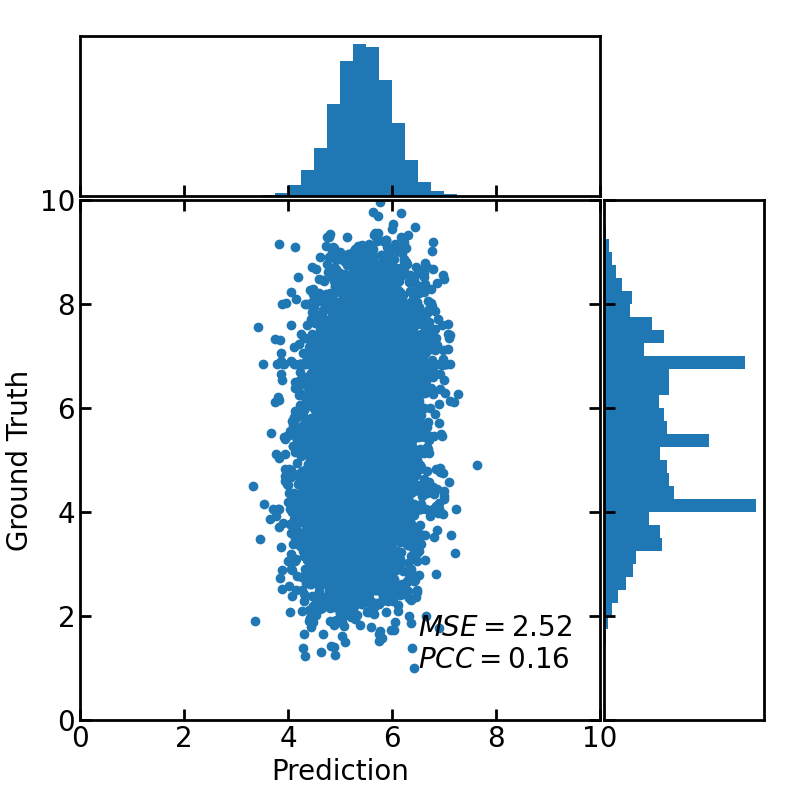}
  \caption{Walkability pre-trained on beauty perception}
  \label{hist:walk}
\end{subfigure}}
\begin{minipage}[b]{0.5\linewidth}
  \begin{subfigure}[b]{0.32\linewidth}
    \includegraphics[width=1\textwidth]{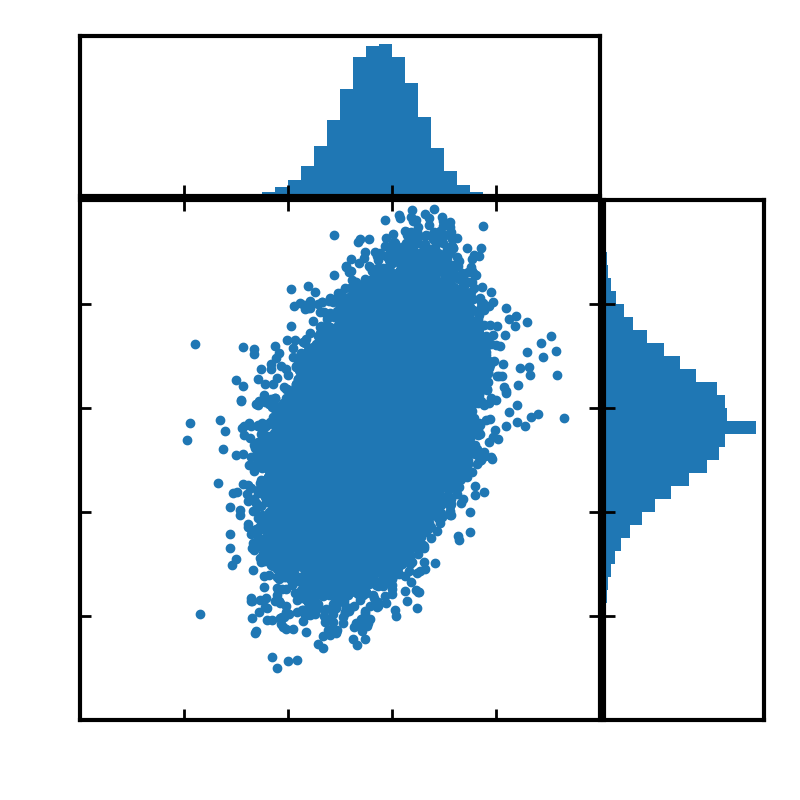}
    \caption{Safety}
    \label{hist:safety}
  \end{subfigure}
  \begin{subfigure}[b]{0.32\linewidth}
    \includegraphics[width=1\textwidth]{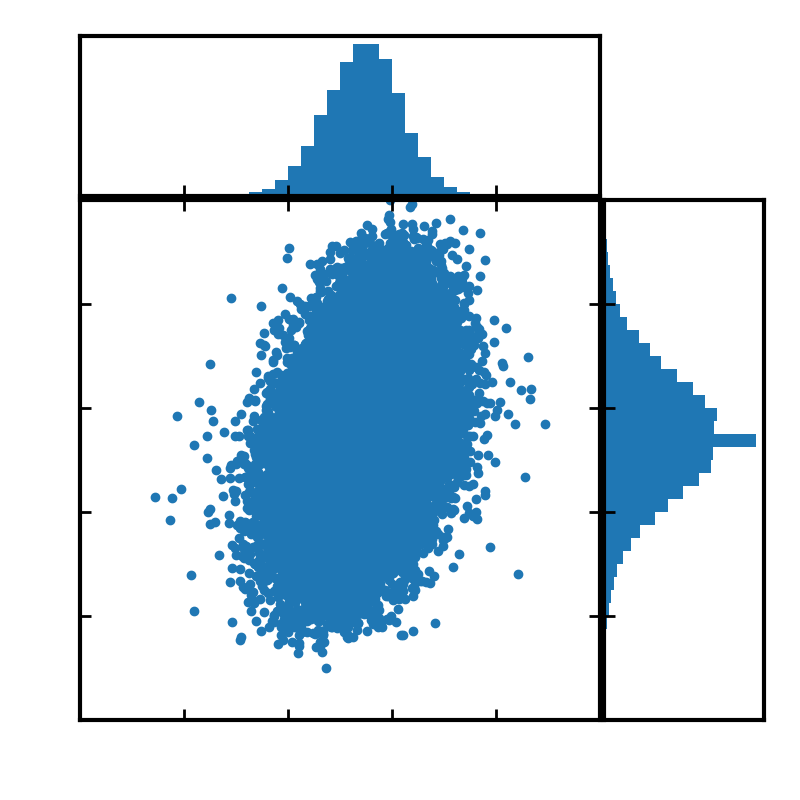}
    \caption{Lively}
    \label{hist:lively}
  \end{subfigure}
   \begin{subfigure}[b]{0.32\linewidth}
    \includegraphics[width=1\textwidth]{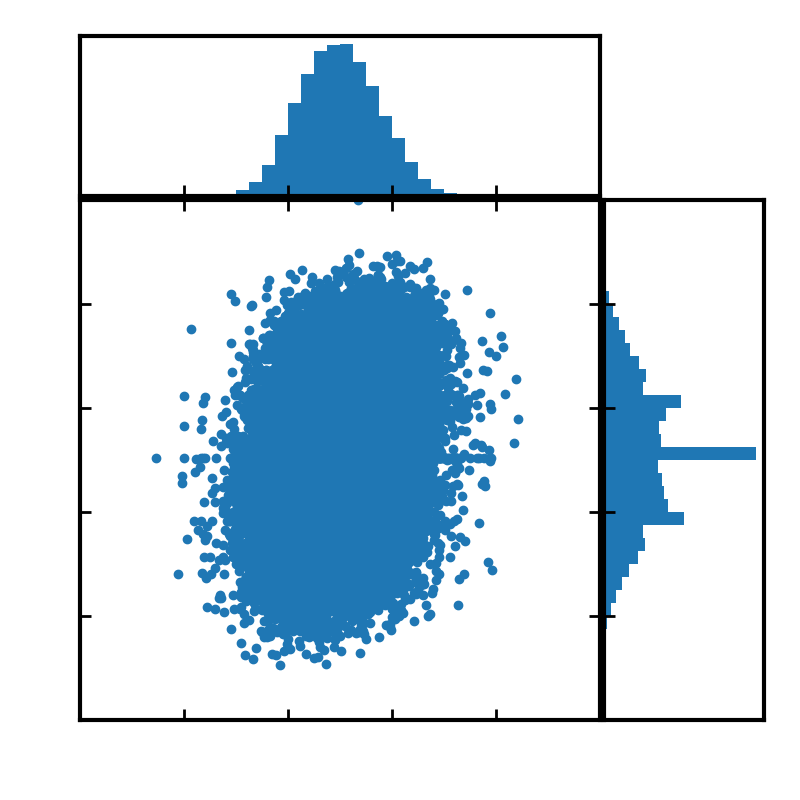}
    \caption{Depres.}
    \label{hist:depressing}
  \end{subfigure}
  \\
  \begin{subfigure}[b]{0.32\linewidth}
    \includegraphics[width=1\textwidth]{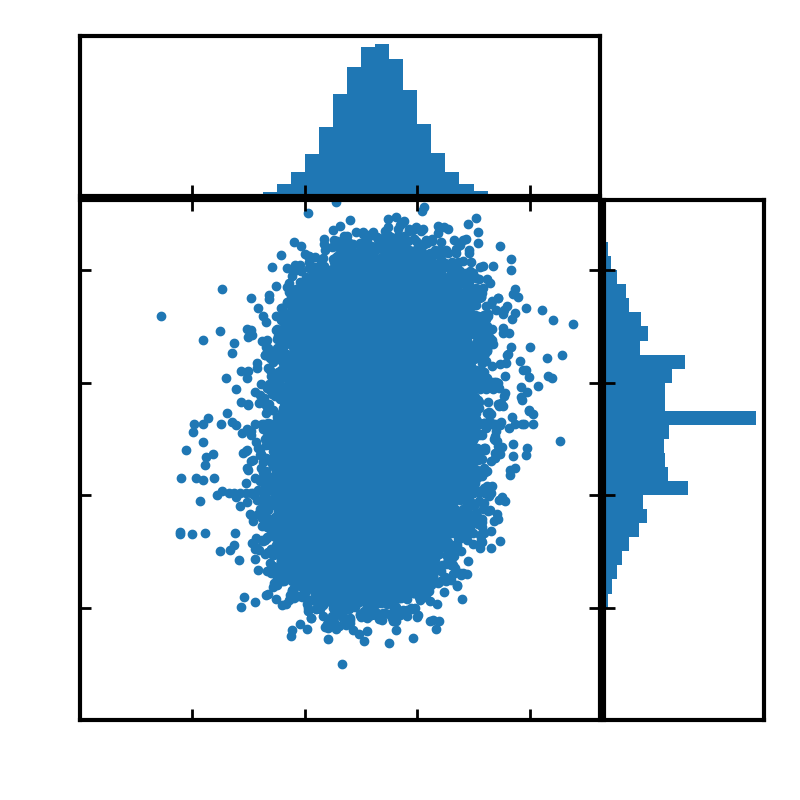}
    \caption{Boring}
    \label{hist:boring}
  \end{subfigure}
   \begin{subfigure}[b]{0.32\linewidth}
    \includegraphics[width=1\textwidth]{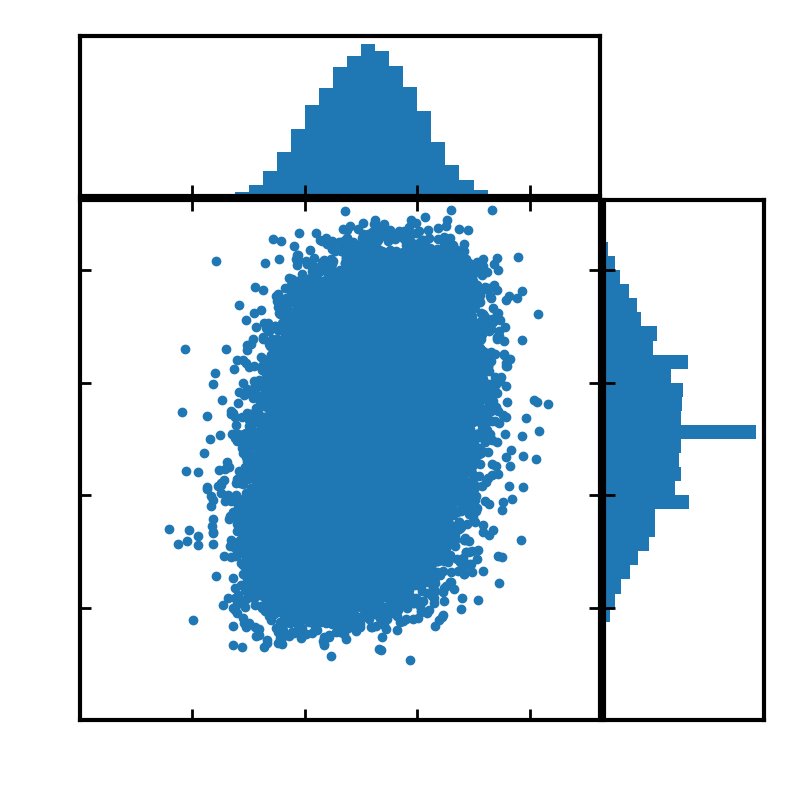}
    \caption{Beauty}
    \label{hist:beauty}
  \end{subfigure}
  \begin{subfigure}[b]{0.32\linewidth}
    \includegraphics[width=1\textwidth]{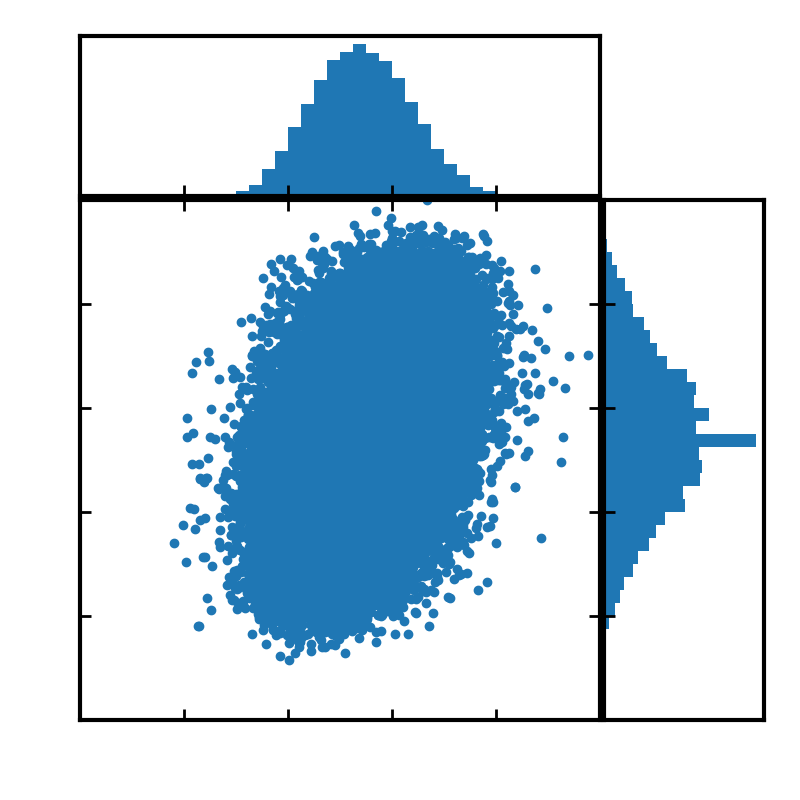}
    \caption{Wealth}
    \label{hist:wealth}
  \end{subfigure}
  \\
\end{minipage}
\caption{Test set prediction versus ground truth scatter plots for original place pulse perceptions and perception of London walkability using Resnet101.}
\label{fig:scatter}
\Description{7 plots for each perception score showing a scatterplot of test set predictions versus ground truth. Plot includes histrogram for both distributions as well as the MSE and PCC scores printed on the plot.}
\end{figure}

For all models, we separately trained a Resnet-101 model to predict perception scores using mean squared error (MSE) loss function. These models were pre-trained on ImageNet \cite{imagenet_cvpr09}. For the Adam optimizer, the initial learning rate was set to $1e-3$ and allowed to decay over $16$ epochs. We scaled the perception scores between $0$ and $10$ and oversampled the images such that each decile was represented equally. For the walkability perception, we compared fine-tuning the ImageNet pre-trained Resnet-101 model, as described previously, with the pre-trained Place Pulse models. The pre-trained beauty model achieves the highest accuracy when fine-tuned on walkability perception (see Table \ref{tab:cnn_results}). Prediction scatter plots are shown in Figure \ref{fig:scatter}. We experimented with freezing the pre-trained network to only perform parameter updates on the final fully connected layers, and performed training on a lower capacity Resnet-18 model pre-trained on ImageNet only. The resulting test set MSE's were $2.52$ and $2.72$ respectively, with Pearson's correlation coefficients (PCC) of $0.15$ and $0.16$ between the predicted and true walkability scores. The low MSE of the frozen Resnet-101 model does not improve the correlation. We therefore used the pre-trained beauty Resnet-101 model for inference on the entire London dataset since it optimises both loss and correlation.

The relationship between the number of games played per image (i.e. the Games Multiplier=No. Games / No. Images), the variance of the predicted TrueSkill scores and the model accuracy and correlation are shown in Table \ref{tab:multiplier}. Safety predictions have the strongest correlation with the ground truth safety perception scores. Safety also has the largest number of games played, resulting in the lowest average variance for image scores, $\bar{\sigma}$. This is not true for the perception wealth, which, although has a lower Games Multiplier, achieves a relatively strong correlation, albeit low CNN test accuracy. This is different from the boring perception, which has slightly improved MSE and Games multiplier and a large reduction in correlation. In Figure \ref{fig:scatter}, we observe that the deep model for wealth perception has done better to capture the variance of the test distribution with little degradation of the MSE. This could indicate a more easily separable ground truth on what makes an image more or less wealthy, compared to what makes an image more or less boring. The performance of each model is a function of the number of games played per image, the variance associated with the ranked scores, and the ability of the network to disentangle features which have been perceived as high or low for a given perception score. For example, the presence of greenery may indicate a high score for walkability, whereas, greenery with no pavement may indicate a low score. In Section \ref{sec:int}, we examine the features which are predictive of high and low perception scores. 

\subsection{Inference}
\begin{figure}[!ht]
\begin{subfigure}[b]{0.9\linewidth}
  \includegraphics[width=1\textwidth, trim=0 20 0 20,clip]{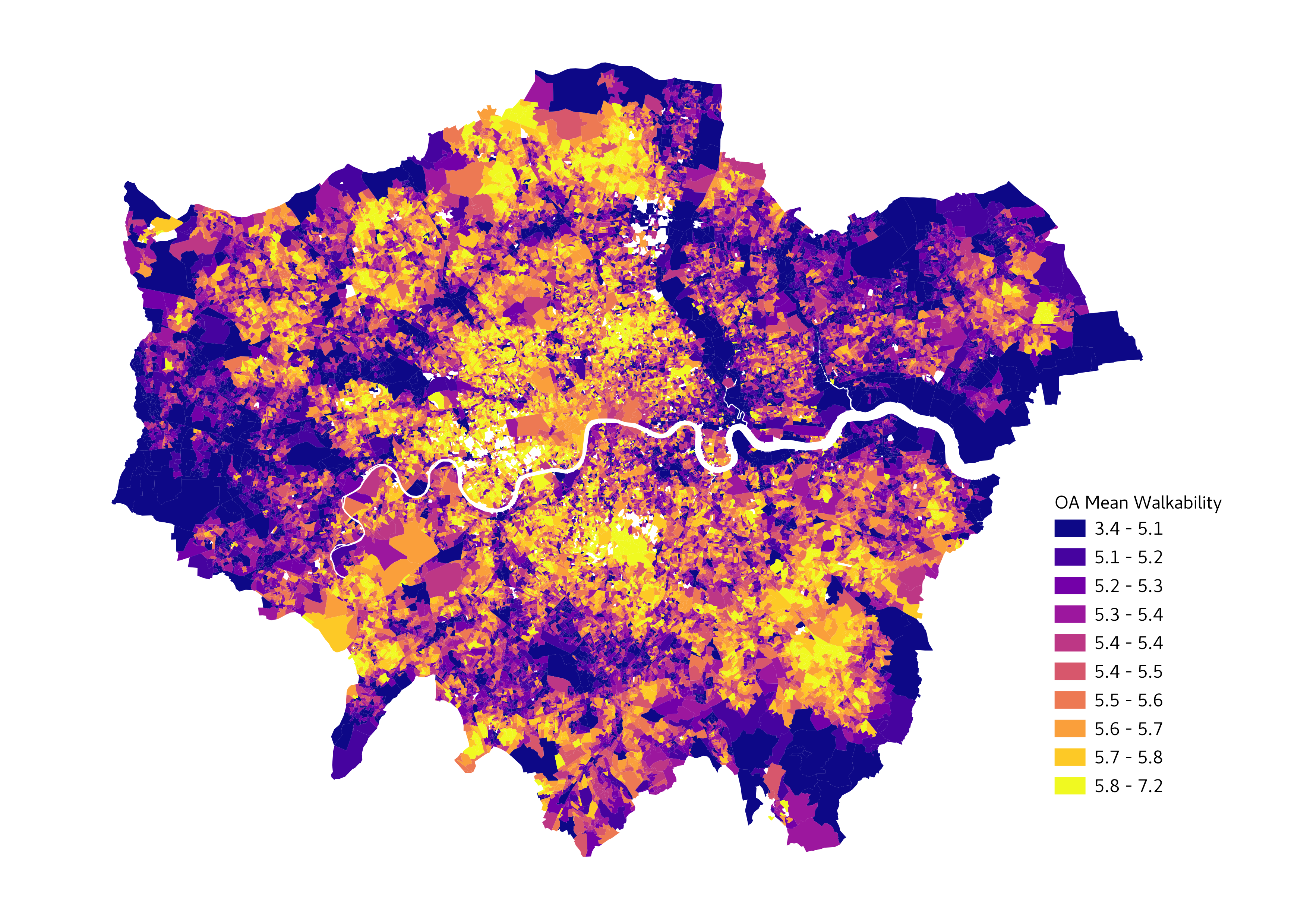}
  \caption{Walkability}
  \label{map:walk}
\end{subfigure}
\begin{minipage}[b]{1\linewidth}
  \begin{subfigure}[b]{0.32\linewidth}
    \includegraphics[width=1\textwidth]{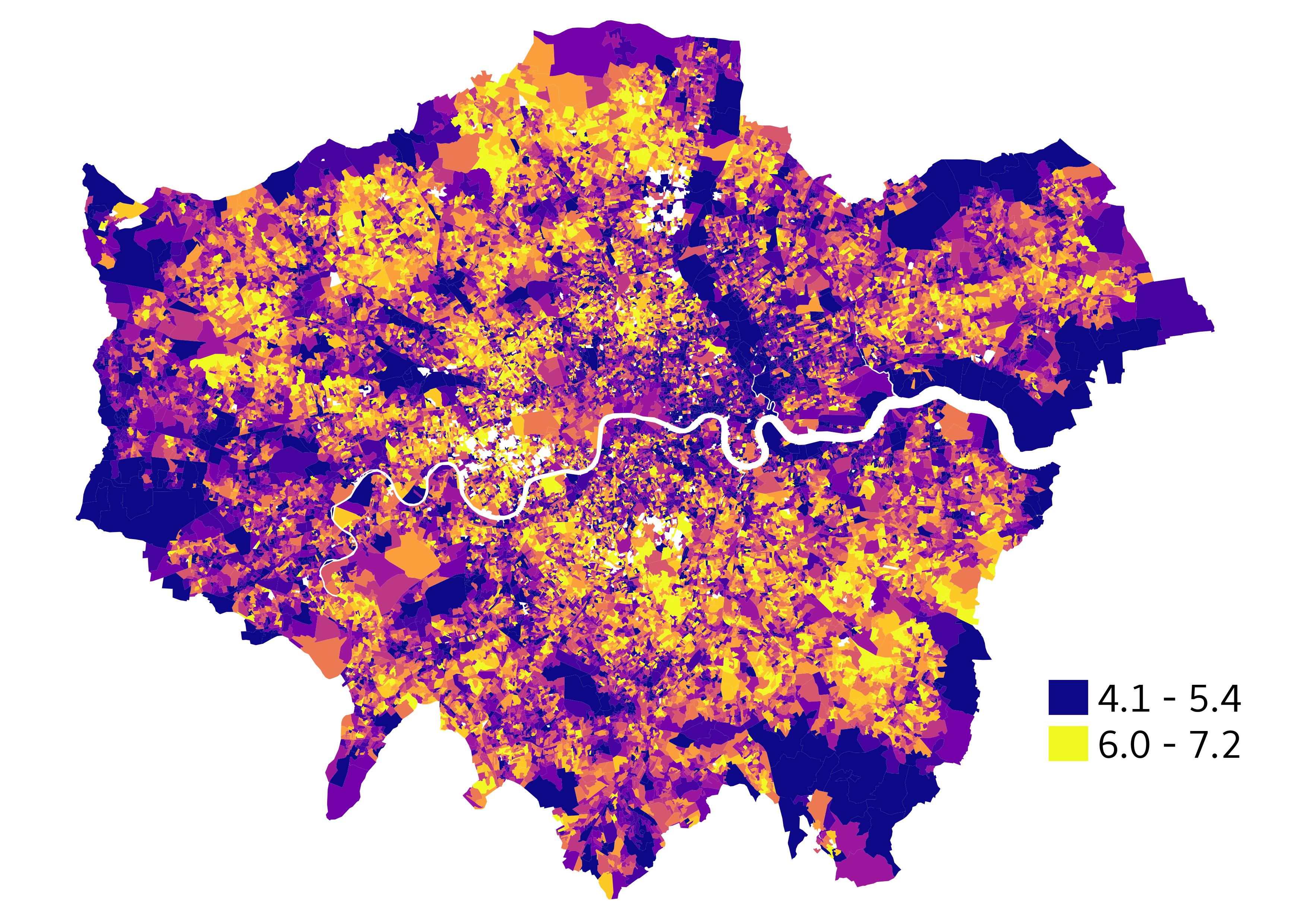}
    \caption{Safety}
  \label{map:safety}
  \end{subfigure}
  \begin{subfigure}[b]{0.32\linewidth}
    \includegraphics[width=1\textwidth]{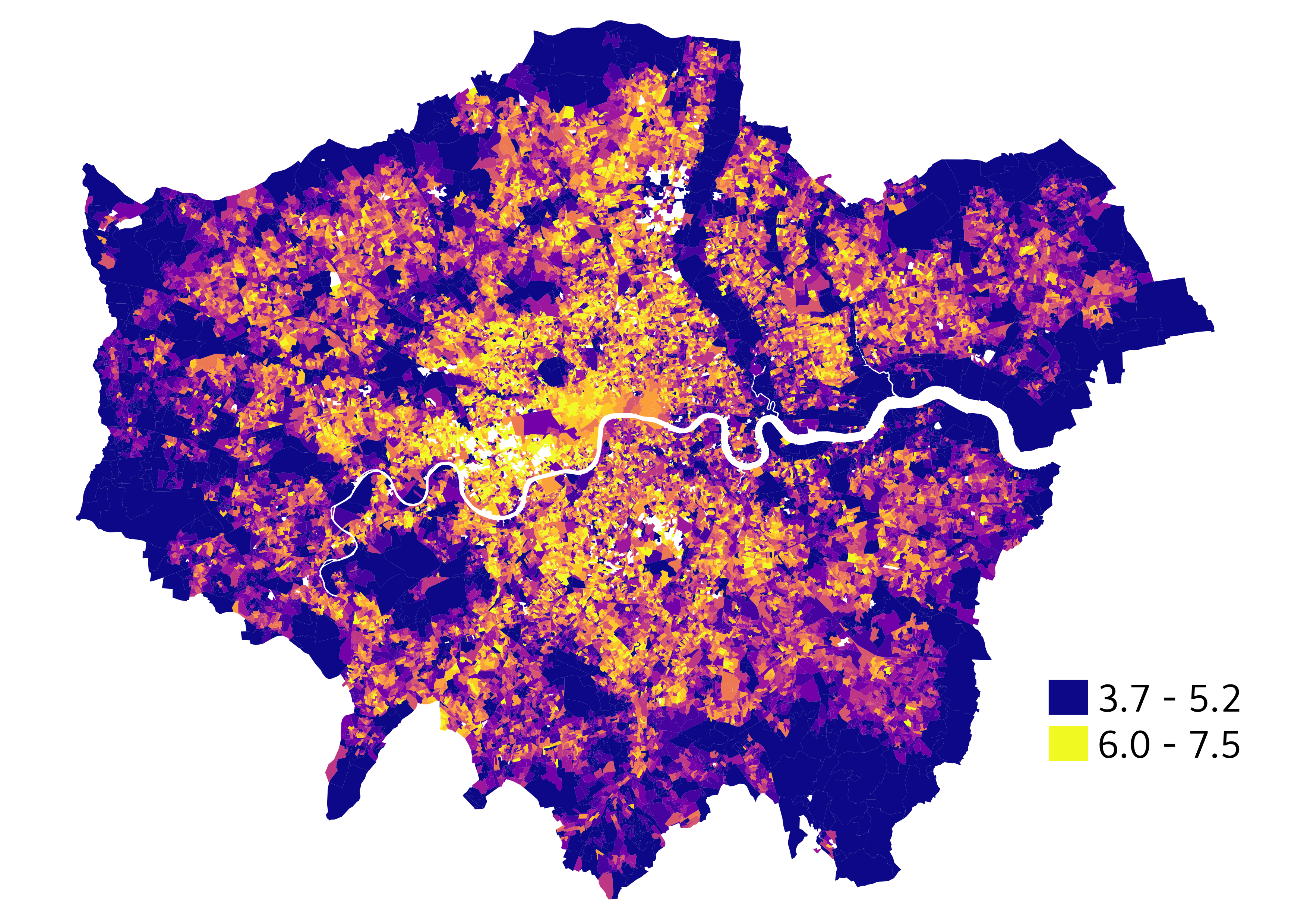}
    \caption{Lively}
  \label{map:lively}
  \end{subfigure}
   \begin{subfigure}[b]{0.32\linewidth}
    \includegraphics[width=1\textwidth]{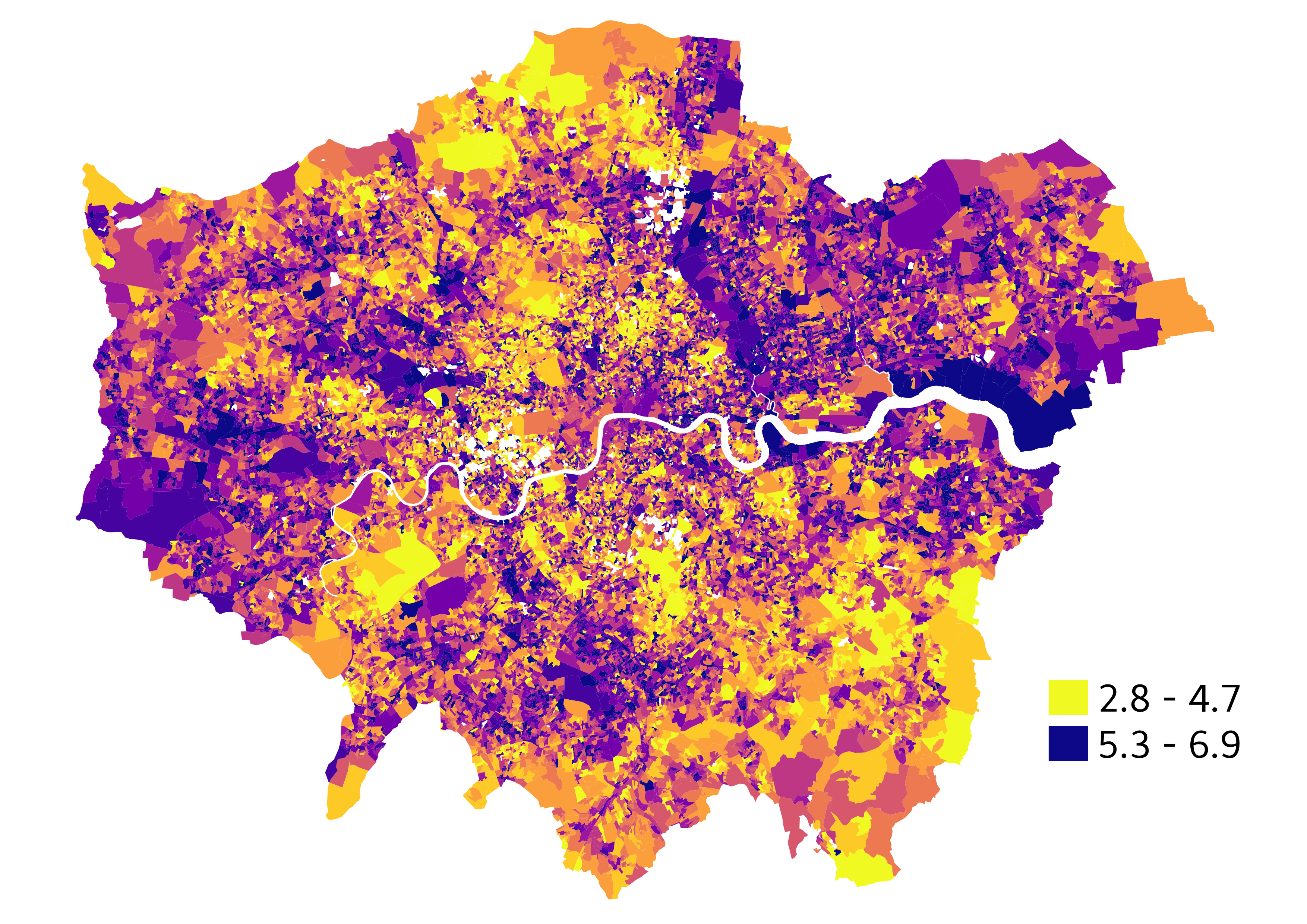}
    \caption{Depressing}
  \label{map:depressing}
  \end{subfigure}
  \\
  \begin{subfigure}[b]{0.32\linewidth}
    \includegraphics[width=1\textwidth]{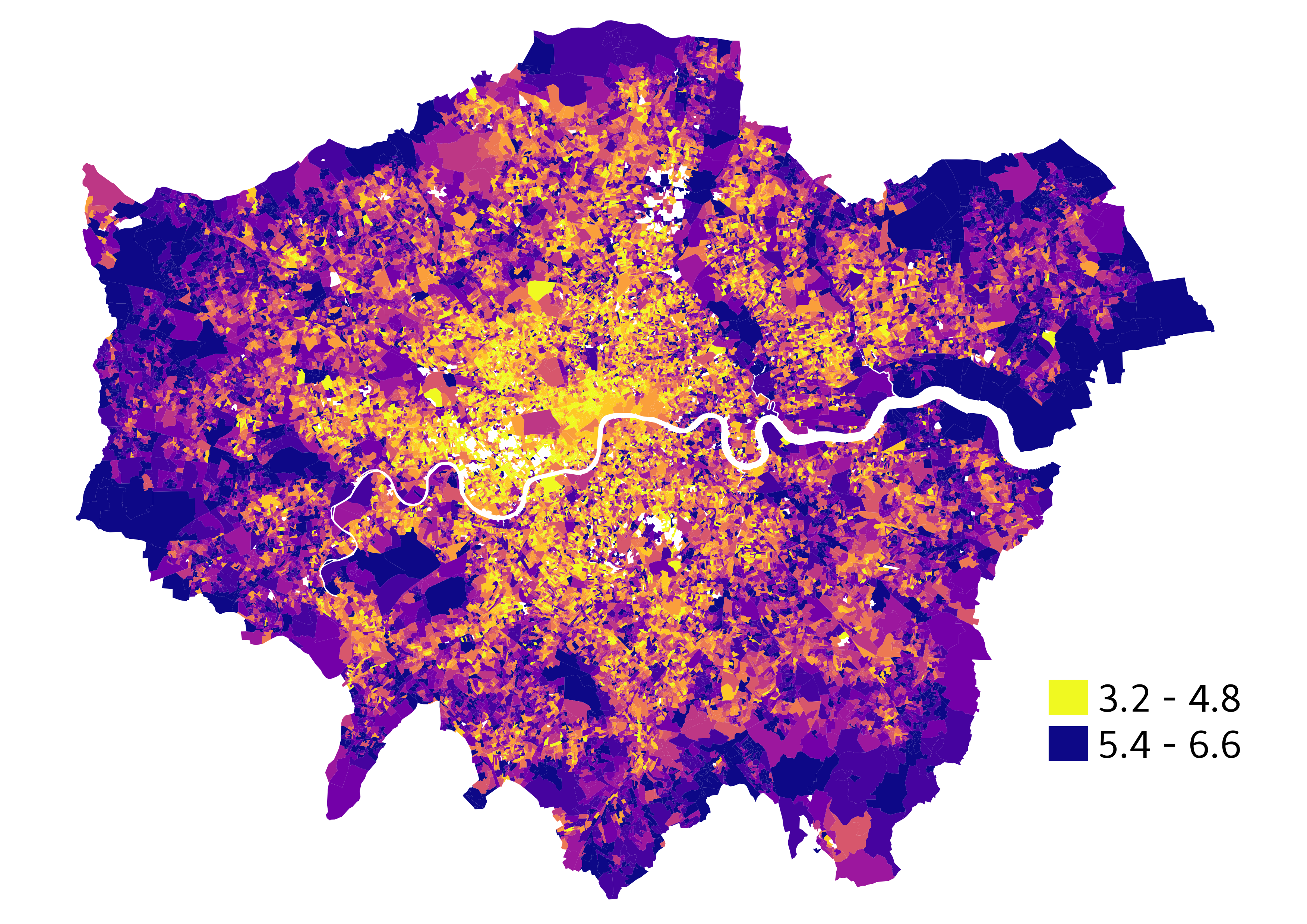}
    \caption{Boring}
  \label{map:boring}
  \end{subfigure}
   \begin{subfigure}[b]{0.32\linewidth}
    \includegraphics[width=1\textwidth]{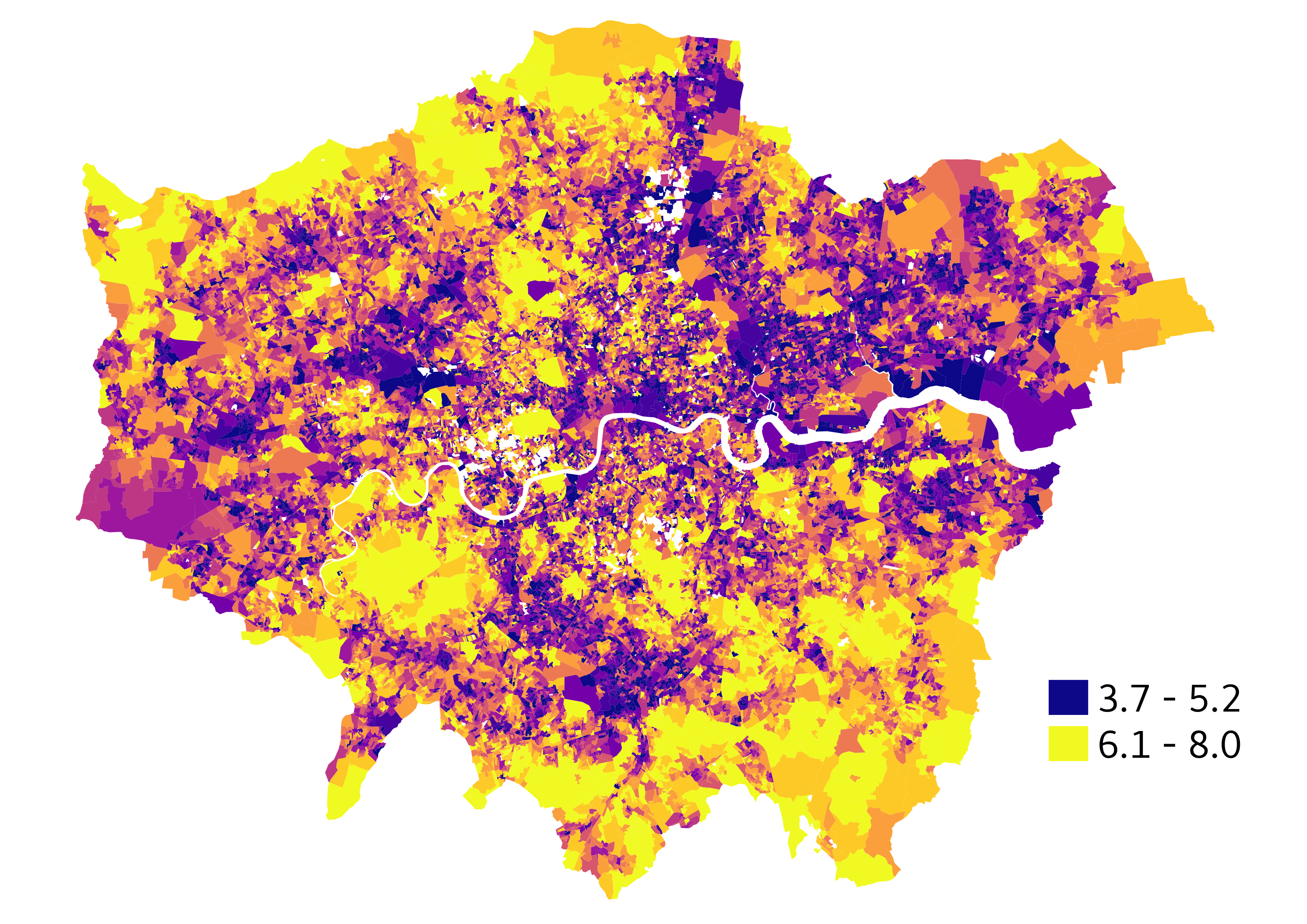}
    \caption{Beauty}
  \label{map:beauty}
  \end{subfigure}
  \begin{subfigure}[b]{0.32\linewidth}
    \includegraphics[width=1\textwidth]{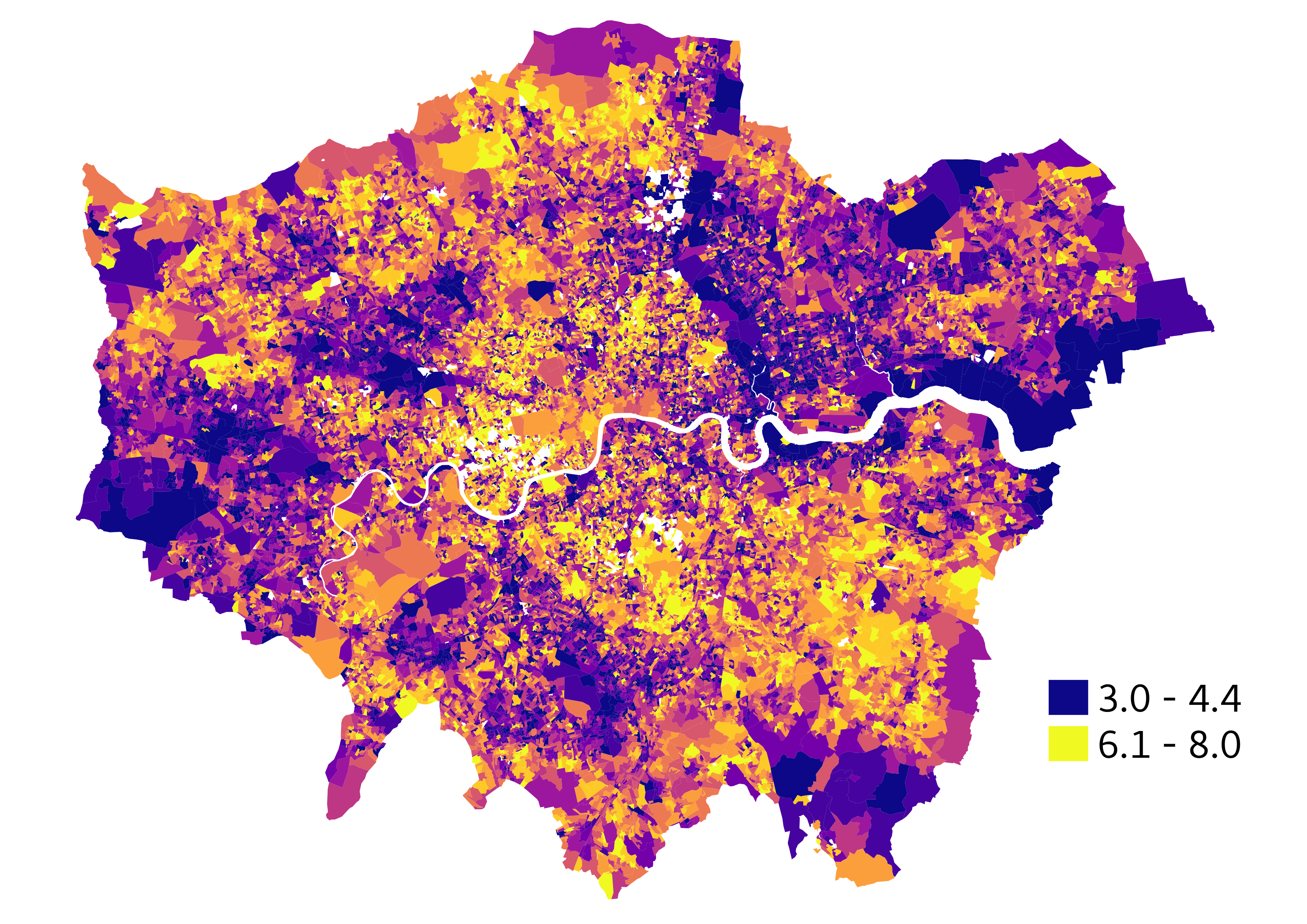}
    \caption{Wealth}
  \label{map:wealth}
  \end{subfigure}
  \\
\end{minipage}
\caption{Deciles of perception scores averaged over Output Areas in London.}
\label{maps}
\Description{Maps from London showing deciles of perception scores at Output Area level for each of 7 perceptions.}
\end{figure}

Using the pre-trained beauty Resnet-101 model fine-tuned on perception of walkability, we performed walkability inference on the 2018 London image dataset containing a total of $1.2$ million images. We averaged the perceived walkability score for all images in an Output Area (OA: the smallest administrative unit in London $\sim 125$ households, median 22 images per OA) and mapped deciles across the entire Greater London extent. In addition, we performed inference using the same image data set using the models trained on the 6 Place Pulse perceptions. All $7$ maps are shown in Figure \ref{maps}. Each perception shows a distinct pattern, suggesting that different features were used to predict each perception. The lowest areas of walkability are found towards the West and East including areas such as Southall, Brentford Hounslow, Harrow and Wembley on the West. Neighbouring districts with higher walkability are Edgware, Northolt and Ealing. On the East, smaller areas of Upminster, Upton and Eastham interrupt the lower perceived walkable areas. On the south side, Croydon and its neighbouring northern suburbs are generally perceived with low walkability, while Crystal Palace, Dulwich and Petts Wood are on the top end of the scale. Large areas of Central London are perceived as walkable and Enfield, Stoke Newington, Kilburn and West Hampstead stand out as highest in the North London area.

\begin{table}[h!]
\centering
\tiny  
\caption{Image level correlations (Pearson) between inferred perception scores for 2018 London images.}
\begin{tabular}{lccccccc}
\toprule
Perception   & Lively	&Boring&	Wealth&	Depressing&	Beauty&	Safety & Walkability \\ \midrule
Lively&	1.00&	-0.14&	0.18&	-0.09&	0.04&	0.23 & 0.20\\
Boring&	-0.14&	1.00&	-0.06&	0.05&	-0.03&	-0.08 & -0.10\\
Wealth&	0.18&	-0.06&	1.00&	-0.13&	0.17&	0.21 & 0.14\\
Depressing &	-0.09&	0.05&	-0.13&	1.00&	-0.15&	-0.14 & -0.11\\
Beauty&	0.04&	-0.03&	0.17&	-0.15&	1.00&	0.14 & 0.14\\
Safety&	0.23&	-0.08&	0.21&	-0.14&	0.14&	1.00 & 0.18\\
Walkability & 0.20 & -0.10 & 0.14 & -0.11 & 0.14 & 0.18 & 1.00 \\
\bottomrule
\end{tabular}
\label{tab:correlations_pp_london}
\Description{Table showing Pearsons correlations over all images for each perception. Diagonal is 1.}
\end{table}

Although the pre-trained beauty model reduced the test error for walkability, liveliness has the greatest correlation, $0.20$ (see Table \ref{tab:correlations_pp_london}). Much of what is rated as the least lively in outer London (note the change in colour scale) is also considered the least walkable. This is not true for beauty, which does have beautifully rated areas in outer London. This suggests the features which are considered more beautiful are more closely related to those which are also considered more walkable. Whereas, liveliness and walkability are related in the co-occurence of objects which are unrelated to the model decision making. For example, the presence of cars might be closely related to the presence of sidewalks. In the next section, we explored feature interpretability for high and low perception scores.

\subsection{Interpretability}
\label{sec:int}
Indepedent from training the CNN to predict perception scores, we extracted features from images using Tensorflow DeepLab API \cite{tensorflow} to obtain object detections and pixel-wise segmentation labels. For segmentation, we used the Xception71 network architecture pre-trained using the cityscapes dataset \cite{mcordts} to predict pixel-wise semantic segmentations on the 2018 London image dataset. For object detection, we inferred bounding boxes using the CenterNet HourGlass 104 model trained of MSCOCO \cite{lin2014microsoft}. There are a total of 90 features in the MSCOCO dataset for object detections and 20 classifications for semantic pixel-wise segmentations. We chose a subset of 21 features from this set which have non-zero feature importance after training a Random Forest classifier across all perception scores. We then trained a Random Forest and Logistic Regression model to predict top and bottom perception scores using the subset of features, performing 5-fold cross-validation and report the average accuracy across all splits (Table \ref{tab:acc}). The classification model for predicting beauty is most accurate, with $>80\%$ accuracy for both models. In addition, there is a small difference between the Random Forest and Logistic Regression models suggesting that the predictors are linearly separable. This is not the case for walkability, which loses $7$ percentage points when predicted using Logistic Regression. While greater accuracy suggests that a larger proportion of explainable features have been captured in the image, we are most interested in interpreting which features correspond to high and low perception scores. There exist explainable methods for classifiers, such as partial dependence plots \cite{friedman2001greedy} and feature importance \cite{breiman2001random} methods, however, Logistic Regression tractably explains all feature importances and the global direction of coefficients. 

\begin{table}[h!]
\small 
\centering
  \caption{Model accuracies for predicting top and bottom decile perception scores. Random Forest (RF) and Logistic Regression (LR).}
  \label{tab:acc}
  \begin{tabular}{crr}
    \toprule
    Perception & Random Forest & Logistic Regression \\
    \midrule
    Safety & 0.71 & 0.67 \\
    Lively & 0.75 & 0.69 \\
    Depressing & 0.64 & 0.63 \\
    Boring & 0.63 & 0.62 \\ 
    Beauty & 0.81 & 0.80 \\
    Wealth & 0.68 & 0.69 \\
    Walkability & 0.74 & 0.67 \\
  \bottomrule
\end{tabular}
\Description{Model accuracies are reported for Random Forest and Logistic Regression classification models (columns) for predicting top and bottom decile for each perception (rows).}
\end{table}


\begin{figure}[ht]
\centering 
    \begin{subfigure}{1\linewidth}
    \centering
  \includegraphics[width=0.65\textwidth, trim=0 0 0 56,clip]{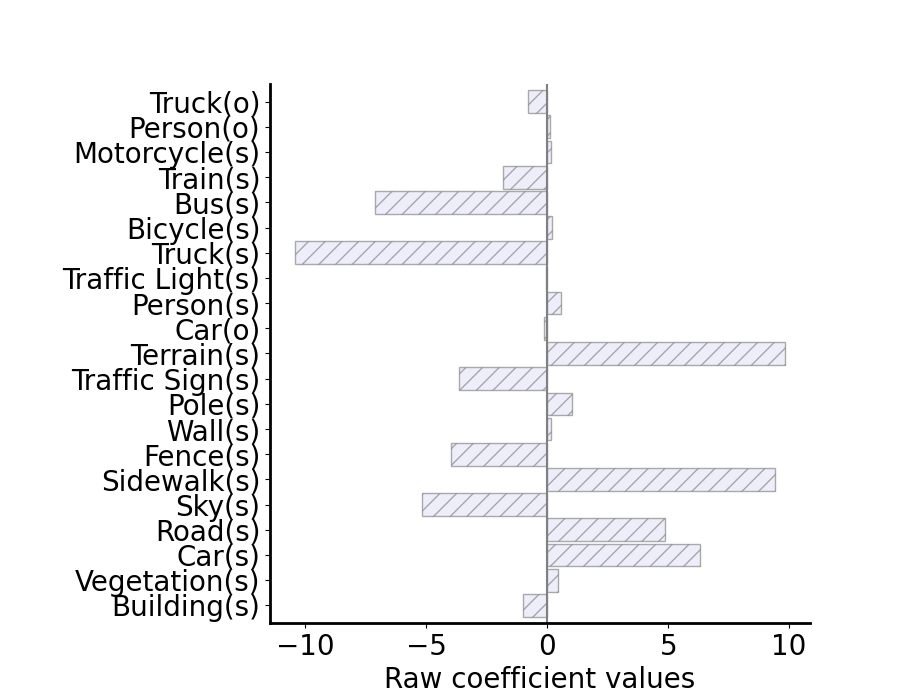}
  \caption{Walkability}
  \label{coef:walk}
\end{subfigure}
\hfill
\begin{minipage}[b]{1\linewidth}
  \begin{subfigure}[b]{0.32\linewidth}
    \includegraphics[width=1.13\textwidth, trim=20 20 20 60,clip]{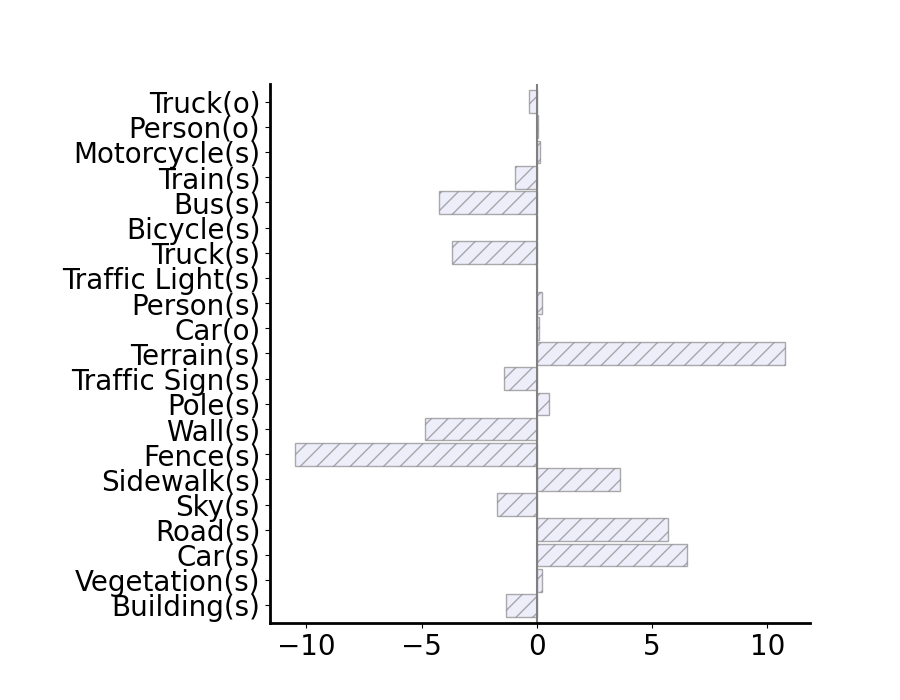}
    \caption{Safety}
  \label{coef:safety}
  \end{subfigure}
  \begin{subfigure}[b]{0.32\linewidth}
    \includegraphics[width=1.13\textwidth, trim=20 20 20 60,clip]{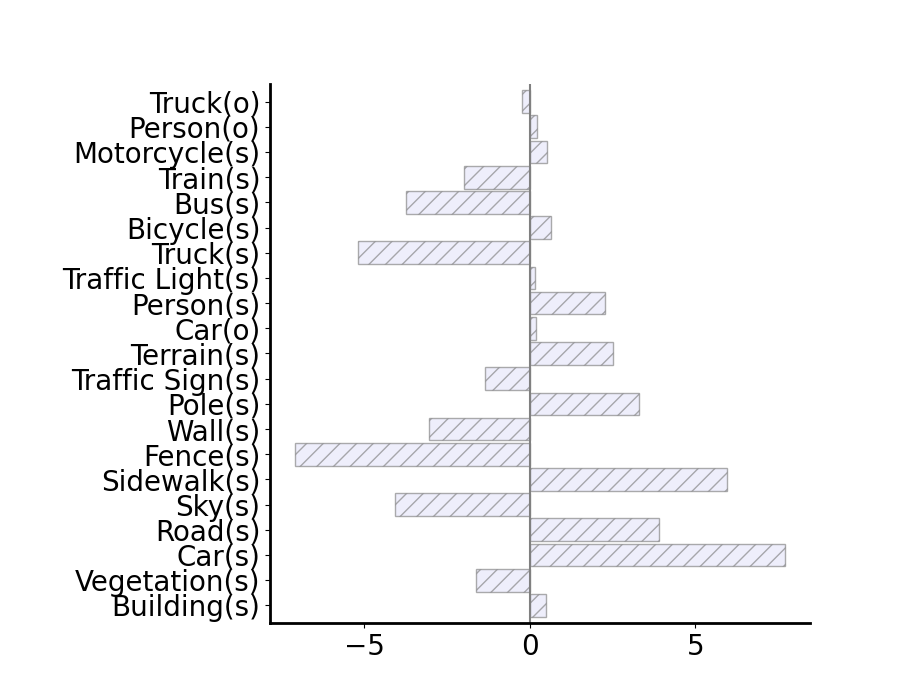}
    \caption{Lively}
  \label{coef:lively}
  \end{subfigure}
   \begin{subfigure}[b]{0.32\linewidth}
    \includegraphics[width=1.13\textwidth, trim=20 20 20 60,clip]{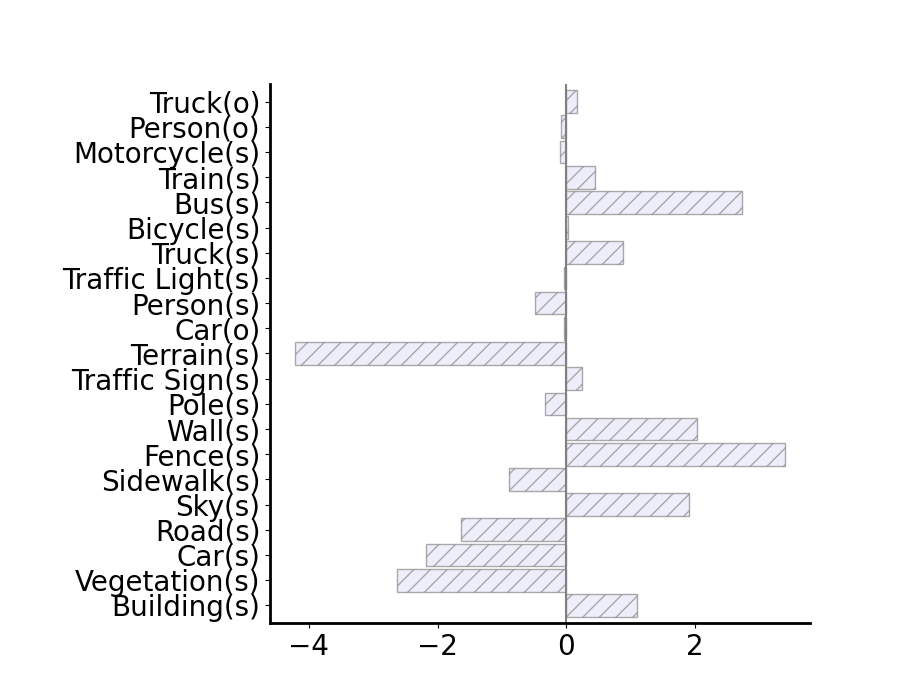}
    \caption{Depressing}
  \label{coef:depressing}
  \end{subfigure}
  \\
  \begin{subfigure}[b]{0.32\linewidth}
    \includegraphics[width=1.13\textwidth, trim=20 20 20 60,clip]{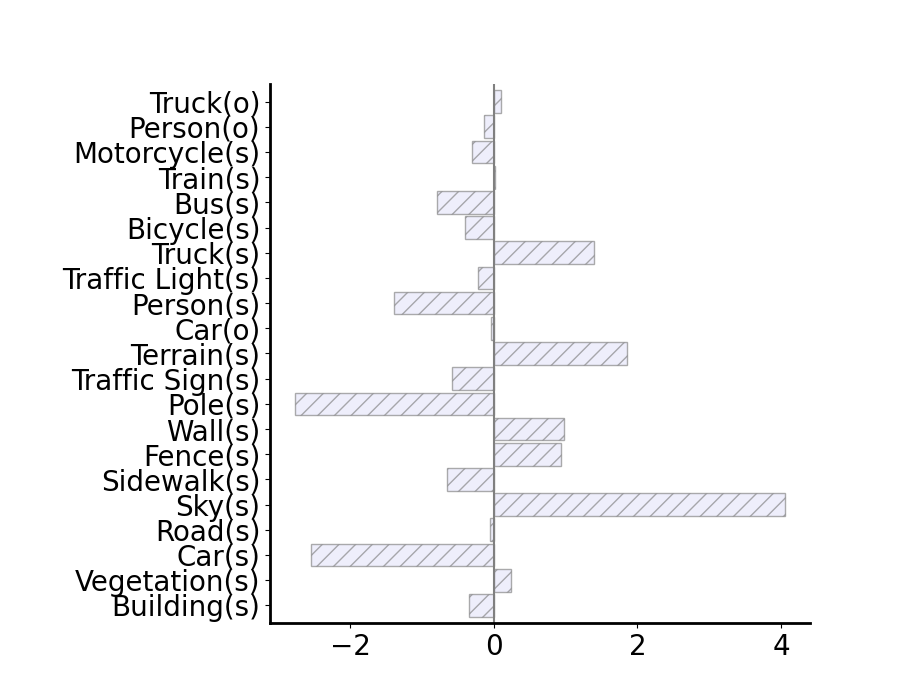}
    \caption{Boring}
  \label{coef:boring}
  \end{subfigure}
   \begin{subfigure}[b]{0.32\linewidth}
    \includegraphics[width=1.13\textwidth, trim=20 20 20 60,clip]{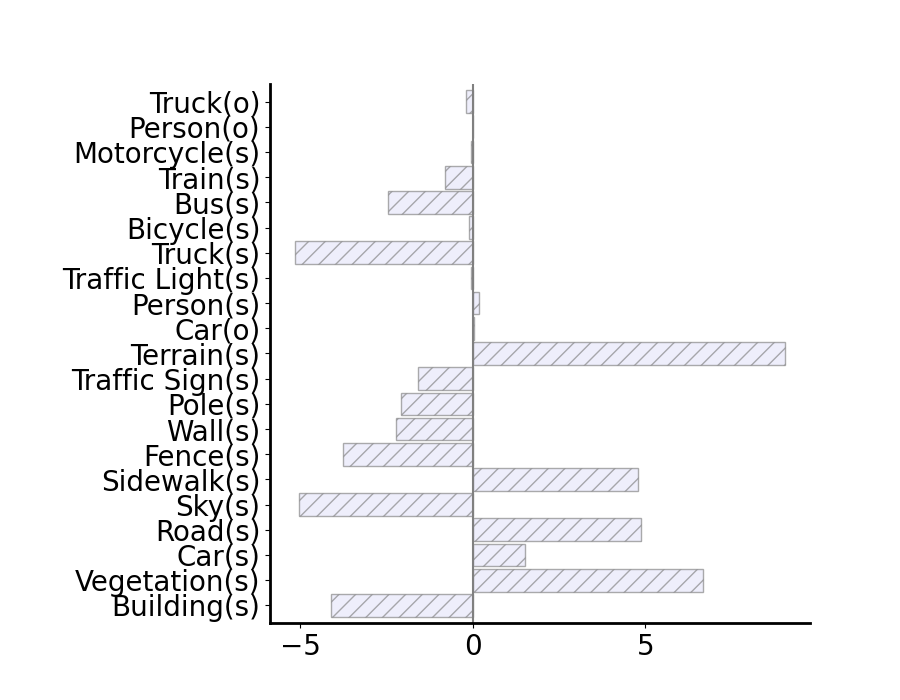}
    \caption{Beauty}
  \label{coef:beauty}
  \end{subfigure}
  \begin{subfigure}[b]{0.32\linewidth}
    \includegraphics[width=1.13\textwidth, trim=20 20 20 60,clip]{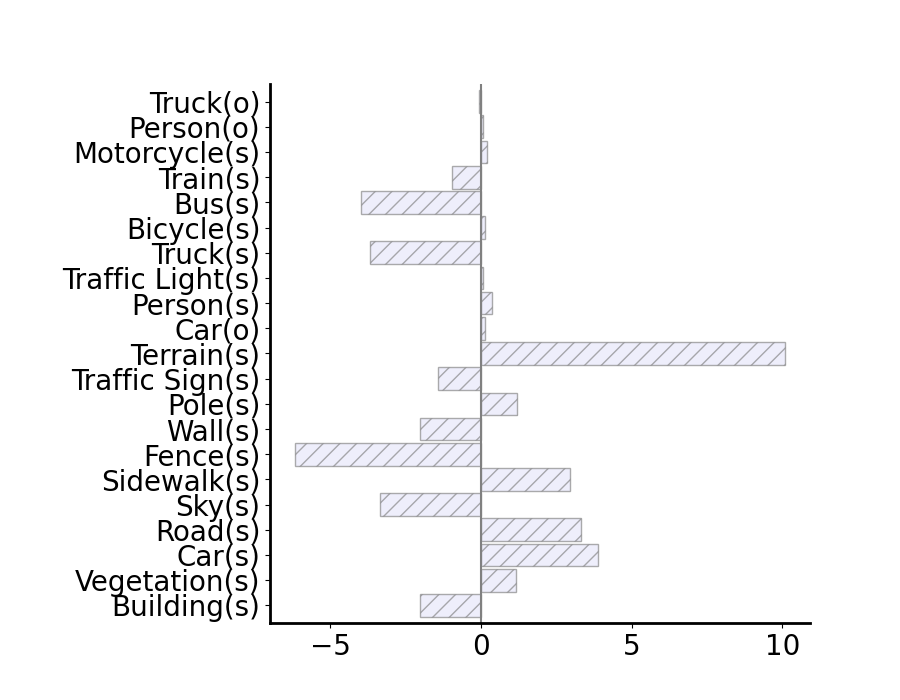}
    \caption{Wealth}
  \label{coef:wealth}
  \end{subfigure}
  \\
\end{minipage}
\caption{Coefficients of Logistic Regression model for predicting top and bottom deciles for each perception score.}
\label{fig:features}
\Description{Bar plots showing the direction and magnitude (x-axis) of coefficients for each feature (yaxis) from Logistic Regression model for each of the 7 perceptions}
\end{figure}

In Figure \ref{fig:features}, the Logistic Regression coefficients are shown to provide a layer of interpretability for the features which contribute to high or low perception scores. For walkability; terrain, sidewalk, road and car have positive prediction values for high walkability. Truck, bus and sky segmentation contribute to a low walkability scores, as do fence and traffic sign to a lesser extent. The coefficient for sidewalk has the strongest impact on the walkability perception (compared to other perceptions) and it is reasonable that other proxies for walkable neighbourhoods (wealthy, safe, beautiful and lively) become unwalkable with the absence of a sidewalk. The presence of truck and bus are indicative of busier roads and contribute to a lower walkability score. This may well also be true for the presence of a traffic light, despite it offering a thoroughfare for pedestrians. That road and car predict walkability could be related to the co-occurence of sidewalks and with greater points of interest. Fence as a predictor of low walkability is somewhat explained by fence also predicting low safety ($PCC=0.18$), low wealth ($PCC=0.14$) and low beauty ($PCC=0.14$), potential proxies for walkability. The impact of vegetation (trees, shrubs, bushes as opposed to terrain which is grass) appears to positively contribute to perceptions of beauty, and wealth to a lesser extent, whereas vegetation has very little impact on walkability. As mentioned previously, vegetation without access to a sidewalk, or terrain to walk on, may prove unwalkable whereas the beauty perception remains intact. Interactions between variables could be further explored to understand these types of relationship. 

\section{Discussion}
\subsection{Survey Collection}
We have built a web app to survey user perceptions of street view images and made our code readily available. Our sampling approach allowed stratified representation of the urban environment, as opposed to area-level or spatial representation as in \cite{salesses_collaborative_2013, dubey_deep_2016}. We included a weighting coefficient to produce more within group image pairs. We chose this strategy to provide finer detailed perceptions on urban environments which might initially appear similar, but have a gradient of quality associated. This approach requires a greater number of ratings to reach a minimum number of games played per image, as we describe below. However, it can be used as a filtering approach to include or exclude images which are not of interest for a chosen research question.

The number of ratings per image influenced the performance on the ranking algorithm with a higher number of games achieving a lower mean standard deviation, as observed for the Place Pulse data set. The TrueSkill ranking algorithm is designed to accommodate a low number of games compared to Elo \cite{elo1978rating}, for example. However, our results lead us to conclude that, employing these methods, researchers should seek a lower bound for the games multiplier of $3$. We were not able to achieve this due to cost restraints. However, we showed that the deep learning model can leverage pre-trained information from the Place Pulse scores when fine-tuning. It is therefore possible to reduce the size of the dataset. Goodfellow et al. suggest a minimum of 5,000 images per class when training deep models \cite{goodfellow2016deep}. Suel et al. find that when transferring models to predict socio-economic outcomes from London to Greater Manchester, just $10\%$ of the target data set is needed to achieve comparable accuracy \cite{suel2019measuring}. In order to justify the minimum number of images needed to collect a city-specific data set, a transferability analysis should be prepared which tests different data set sizes. Even so, this should be used as a guideline since cities have varying distributions for their image data sets which may require greater representation when model training. For example $40\%$ of the cities represented in the Place Pulse data set are from Europe, while just $3$ African cities are represented \cite{dubey_deep_2016}.

Unlike Place Pulse \cite{dubey_deep_2016}, we collected demographic information from our participants to estimate group differences in preferences. We ensured no personal information was collected. We found significant differences (at a $95\%$ confidence interval) across those individuals who were from London versus not from London and the groups of high versus low activity. These differences were seen across 3 (London groups) and 5 image pairs (activity groups) out of a total of 14 image pairs which were repeated across users, while there were no significant differences between male and female group. Salesses et al. found no significant effects for group differences of age, gender and US versus non-US based locations \cite{salesses_collaborative_2013}. Their approach tests the differences in perception scores when ranked according to each group. Group-based differences may be smoothed over when considering the global image ranking as opposed to individual preferences for image pairs. Our approach allows for testing users group differences by injecting repeated samples across user-games. We randomly selected a subset of images to be repeated. To improve on this strategy, repeated image pairs should be carefully created to represent a broad set of perception decisions that are relevant for the research question.

We disseminated the platform through professional and personal networks to gather perceptions. We then consulted Amazon Mechanical Turk to further collect perception ratings, paying workers above the minimum wage. We were not able to show the differences between AMT/non-AMT since the same repeated image pairs were not delineated for these groups. This was due to an update in the database which meant that the repeated game images were mixed with other images. AMT workers have a low within group agreement score of $56\%$. AMT workers could have greater within group variability compared to participants from convenience sampling through personal networks, who we might assume are more socio-demographically similar. Future work should compare group differences between participation networks to validate AMT as method of data collection for urban perceptions.

\subsection{Model training and Interpretability}
Deep neural networks are capable of approximating arbitrarily complex functions. Leveraging large amounts of data, deep models learn patterns, features, color gradients and the combination of these and more, to predict outcomes. We found the perceptions scores to be weakly correlated, but largely, each model predicts a unique and different set of scores for each image. It can be that deep models learn spurious features which are correlated with the outcome, such as hospital-specific metal tokens in
the chest X-ray scans for predicting pneumonia \cite{zech2018variable}. However, the probability of these spurious correlations over all 7 perceptions is unlikely. It is therefore reasonable to assume that the model has learnt a function for approximating human-based perception scores. We apply model-agnostic interpretability methods to assess features which correspond to high and low perception scores. Model-specific methods for interpretability of CNNs, such as pixel-attribution methods \cite{molnar2022}, assess the models decision making instead of mapping concepts to the outcome class. However, these methods are difficult to benchmark against expectation from a human observer without including user-based validation. Human interpretable features, such as objects and semantic pixel-wise labeling, are therefore more readily available for validation. 

The image scores follow a normal distribution around the initialised $\mu=25$. To provide more information to the CNN about the tails of the distribution, we oversampled images from the tails. Other approaches can use augmentation \cite{shorten2019survey} or adjust the loss function \cite{cui2019class} to manage imbalanced distributions. 

\subsection{Mapping}
The maps of perceptions shown in Figure \ref{maps} differ from previous studies based on coverage, which is higher than other street-view mapping literature: $150$K point locations in London \cite{suel2019measuring} (compared to $630$K used in our study), $50$ meter intervals in Beijing and Shanghai \cite{zhang2018measuring} but only collecting $245$K and $150$K images respectively. We were able to achieve close to $70\%$ street coverage by sampling points on the road at $20$ meter intervals. Additional years can be used to cover a greater proportion of streets, or missing point locations can be gathered by augmenting different sources of data.

These maps could be added to city walk scores and made available online for citizens to have access to an additional layer of information when making decisions on activity or housing, for example.

\subsection{Limitations}
As mentioned in the previous sections, the study requires scaling user-based perceptions ratings which can be costly to acquire. Future work on transferability with lower amounts of data should help to guide thresholds for survey collection. 

The use of crowd-sourcing and the embedded positionality of the survey participants can directly effect the outcome of interest \cite{diaz2022crowdworksheets}. While we do not assume that there is one correct answer in this work, we do assume that the average converges to a meaningful interpretation of images that can be considered more or less walkable over the survey population as in \cite{dubey_deep_2016}. It is important to note, however, that our average may be taken from a skewed socio-demographic population based on preference or restriction to home-based work \cite{ipeirotis2010demographics}, as one example. Our paper is limited by not studying skewed annotator demographics in the crowd-sourced survey population. 

The model accuracy for our newly collected perception of walkability is low, relative to other perceptions. We have discussed in the previous section on how this might be mitigated. Future work on city-wide mapping could include uncertainty estimates in the final spatial maps to reflect the performance of the model on this perception score.


\section{Conclusion}
In this study we have created a methodological toolkit for assessing city-wide perceptions of urban perceptions. This work is intended to equip researchers with a user-based metric for peoples experience of the urban environment and provide a meaningful interpretation for user activity. Future work on temporal analysis can identify the impact of urban interventions in modern cities. This method can be adopted within urban planning to identify proposal regions for urban investment and regeneration. 

\section*{Acknowledgements}
We thank Giulia Mangiameli for project management and coordination of activities. This work was supported by the MRC Centre for Environment and Health PhD Studentship and Pathways to Equitable Healthy Cities grant from the Wellcome Trust [209376/Z/17/Z].

\bibliographystyle{ACM-Reference-Format}
\bibliography{sample-base}


\end{document}